\documentclass[lettersize,journal]{IEEEtran}
\usepackage{algorithm}
\usepackage{algorithmic}
\usepackage{amsmath, amsfonts}
\usepackage{array}
\usepackage{textcomp}
\usepackage{stfloats}
\usepackage{url}
\usepackage{verbatim}
\usepackage{graphicx}
\usepackage{cite}
\usepackage{xcolor}
\usepackage{multirow} 
\usepackage{booktabs}
\usepackage{array}
\usepackage{booktabs}
\usepackage{makecell}

\usepackage{ulem}

\usepackage{multirow}
\usepackage{pifont}
\newcommand{\cmark}{\ding{51}} 
\newcommand{\xmark}{\ding{55}} 

\usepackage{lipsum}
\usepackage{tikz}
\usepackage{graphicx}
\usepackage{caption}
\usepackage{array}
\usepackage{multirow}
\usepackage{adjustbox}
\usepackage{geometry}
\usepackage{balance}
\usepackage{soul}
\usepackage{xcolor}
\usepackage{subcaption}
\usepackage{hyperref}

\usepackage{float}
\begin{document}

\title{MCLC-NET: Multimodal Continual Learning for Leaf Counting}


\author{Ruchi Bhatt, Pratibha Kumari, Shreya Bansal, Vedant Agnihotri, Dwarikanath Mahapatra, Mukesh Saini
\thanks{The authors Ruchi Bhatt, Shreya Bansal, Vedant Agnihotri, and Mukesh Saini are with the Department of Computer Science \& Engineering, Indian Institute of Technology Ropar, Punjab, India (email: ruchi.21csz0007@iitrpr.ac.in; shreya.22csz0010@iitrpr.ac.in; 
09vedantagnihotri1@gmail.com;
mukesh@iitrpr.ac.in), Pratibha Kumari is with the Department of Informatics and Data Science, University of Regensburg,
Regensburg, Germany (pratibha.Kumari@ur.de)   Dwarikanath Mahapatra is with the Department of Computer Science at Khalifa University, UAE (dwarikanath.mahapatra@ku.ac.ae). 

}

}




\maketitle

\begin{abstract}
Leaf counting is an important task in plant phenotyping for monitoring plant growth and estimating crop yield. Most existing methods rely on RGB images, but their performance is often affected by occlusion, lighting variations, and other real-world challenges. Additional modalities, such as depth and thermal images, can provide useful complementary information. However, multimodal leaf counting remains underexplored. Also, many existing methods assume that all training data are available simultaneously, which is impractical in real agricultural settings, where data is collected over time from multiple sources. To address these challenges, we propose MCLC-NET, a multimodal continual learning framework for leaf counting. It learns tasks sequentially using a memory-based strategy with a memory buffer to retain important samples from previous tasks. We also introduce MMLC, a real-world multimodal leaf-counting dataset designed for a domain-incremental scenario (DIS) in CL. It contains RGB, depth, and thermal images collected across different crop types under varying environmental conditions, arranged in three orderings: crop-wise, time-wise, and mixed. Experimental results, averaged over three random seeds, demonstrate that MCLC-NET consistently outperforms existing methods across all three task orderings, achieving the lowest AMSE of 0.675$\pm$0.027, 0.542$\pm$0.069, and 0.745$\pm$0.057, respectively. The MMLC dataset is available at
\href{https://drive.google.com/drive/folders/1\_tMjExi5yjgtiaj-wZSP2qtFnElwqXHL}{MMLC Dataset}.


\end{abstract}


\section{Introduction} \label{sec:intro}


One of the key tasks in agricultural research is plant phenotyping~\cite{buzzy2020real,li2020review}, which involves analyzing visible plant traits to monitor growth and development. Among these traits, leaf counting~\cite{quamer2026twoCOMPAG}
is particularly important, as it provides valuable insights into the plant's health~\cite{minervini2016finely,chen2020usingCOMPAG}, growth stage~\cite{buzzy2020real,yue2020predictionCOMPAG}, and expected yield potential~\cite{dobrescu2017leveraging}\cite{sarkarTAFE}. 
Traditionally, researchers have developed image-based methods to automate leaf counting that rely solely on unimodal images, typically RGB images. While RGB images are easily captured and widely used, they often face difficulties in real agricultural conditions. Problems such as occlusion, varying lighting, and complex backgrounds can affect their performance in real-world conditions~\cite{giuffrida2018pheno}. 
As a result, models trained only on RGB data may not generalize well in practical scenarios. To improve this, researchers have begun exploring multimodal approaches that combine RGB data with other modalities, such as depth and thermal images.

Each modality provides different information. RGB captures color and texture, thermal images reflect temperature-related patterns, and depth images provide information about relative object distance. When used together, these modalities can complement each other and help the model better handle challenging conditions, such as poor lighting or occluded leaves. Despite its vast potential, only a few studies have explored leaf counting using multimodal data~\cite{giuffrida2018pheno}, making this an underexplored area. Also, existing multimodal datasets suitable for leaf counting, such as MSU-PID \cite{cruz2016multi}, are limited in number and are primarily collected in controlled environments. These settings may not reflect the variability and complexity of real field conditions. To address this limitation, we collect a new dataset from real farms comprising RGB, depth, and thermal images of different crops under varying conditions.




Most existing leaf-counting methods rely on one-time training and deployment of deep models and assume that all training data is available at once. In practice, this assumption does not hold, as in real deployments, data is not available all at once. Instead, it is collected sequentially over time across different sources, seasons, locations, and environmental conditions. This makes the retraining from scratch impractical due to computational and storage constraints. This creates a need for models that can continuously adapt to new data distributions while retaining previously learned knowledge. 
However, a key challenge in such settings is that models tend to forget previously learned information when trained on new data. This issue is known as "catastrophic forgetting" ~\cite{Kum_Continual_MICCAI2024}.
Because of this, standard deep models are not well-suited to real-world agricultural applications where conditions are constantly changing. CL~\cite{kirkpatrick2017overcoming,lopez2017gradient} addresses this problem by allowing models to learn sequentially from new data while retaining past knowledge. This approach allows the model to adapt to changing environments, such as new crop types, growth stages, or imaging conditions, without losing performance on older data. We formulate leaf counting as a domain-incremental scenario (DIS) problem \cite{mirza2022efficient}, where the task remains the same but the data distribution changes across domains such as crop type, capture time, and environmental conditions. Unlike class-incremental learning \cite{tao2020few}, the goal is not to learn new categories but to adapt to these distribution shifts while preventing catastrophic forgetting. 

Among CL strategies, namely regularization~\cite{kirkpatrick2017overcoming}, architectural~\cite{yoon2017lifelong}, and rehearsal-based (memory-based)~\cite{lopez2017gradient}, rehearsal-based methods are widely used due to their simplicity and effectiveness. According to the literature, generative replay avoids storing past data but requires complex sample generation~\cite{shin2017continual}. In contrast, memory-based replay is simpler, often more effective, and widely used in CL \cite{bidaki2025online}, which motivates our choice in this work.
Although rehearsal-based CL methods have shown promising performance, most existing approaches are primarily designed for classification problems and unimodal settings. Their direct application to multimodal leaf counting is challenging due to the regression nature of the task and the need to effectively utilize complementary information from multiple modalities. Moreover, existing memory-based replay strategies often rely on random sampling, which may result in the storage of redundant or less informative samples. In contrast, the proposed framework focuses on multimodal continual regression. It uses uncertainty- and diversity-aware sample selection rather than random sampling to retain representative samples from previous tasks while adapting to new domains.

In this paper, we propose a rehearsal-based framework, Multimodal Continual Learning for Leaf Counting (MCLC), that combines the strengths of multimodal data and CL to provide a robust, adaptive leaf-counting model. MCLC integrates samples from multimodal datasets into a single representation and learns sequentially across tasks using a memory-based CL strategy. We store a few important samples from previous tasks and use them to represent past data while learning new tasks. Here, each task refers to a subset of data, defined by crop type, capture time, and environmental conditions. This DIS setup reflects real-world agricultural scenarios and highlights the need for CL in leaf counting. To support this, we introduce the Multimodal Leaf Counting (MMLC) dataset, collected from real farms. To the best of our knowledge, we are the first to explore multimodal continual learning for domain-incremental leaf counting using RGB, depth, and thermal modalities. 


The main contributions of this paper are highlighted below:

\begin{itemize}

\item  We propose MCLC, a multimodal memory-based CL framework for leaf counting that combines uncertainty-aware and diversity-aware sample selection to retain informative samples from previous tasks while adapting to new tasks.


\item We introduce MMLC, a real-world multimodal leaf counting dataset containing paired RGB, depth, and thermal images collected from different crops under varying environmental conditions for domain-incremental CL.



\end {itemize}







The rest of the paper is structured as follows. Section \ref{sec:literature} contains the related works. Section \ref{sec:dataset} consists of a detailed description of the proposed dataset. Section \ref{sec:method} includes the proposed methodology. Section \ref{sec:experiment_results} consists of the
experiments and results, followed by an ablation study in Section \ref{ablation}. The conclusion
is given in section \ref{conclusion} followed by limitations and future work in section \ref{futurework}. 
\begin{table*}[!t]
\centering
\caption{Comparison of existing methods with our proposed approach. DIL refers to domain-incremental learning.}
\label{tab:comparison_lit_final}
\renewcommand{\arraystretch}{1.3}
\scalebox{0.85}{
\begin{tabular}{llllp{4cm}lll}
\toprule
\textbf{Method} &  \textbf{Regression} & \textbf{CL}  & \textbf{DIL}  & \textbf{Modalities} & \textbf{Application} & \textbf{Field Data} & \textbf{Dataset / Remarks} \\
\midrule
Giuffrida et al.~\cite{giuffrida2016learning} &  \cmark & \xmark & \xmark & RGB & Leaf Count & \cmark & CVPPP \\

Dobrescu et al.~\cite{dobrescu2017leveraging} &   \cmark & \xmark & \xmark & RGB, NIR, Fluorescence, Depth & Leaf Count & \cmark & CVPPP + Custom \\

Giuffrida et al.~\cite{giuffrida2018pheno} &   \cmark & \xmark & \xmark & RGB, NIR, Fluorescence & Leaf Count & \cmark & MSU-PID \\

Xu et al.~\cite{xu2018leaf} &  \xmark & \xmark & \xmark & RGB & Segmentation & \cmark & DeepLab-like \\

Itzhaky et al.~\cite{itzhaky2018leaf}  & \xmark & \xmark & \xmark & RGB & Density Map & \cmark & Leaf segmentation \\

Tu et al.~\cite{tu2020automatic}  & \xmark & \xmark & \xmark & RGB & Detection & \cmark & YOLO-based \\

Sun et al.~\cite{sun2020multimodal} &  \xmark & \cmark & \xmark & RGB, Text, Audio &  Classification & \xmark & Modality dropout \\

Srinivasan et al.~\cite{srinivasan2022climb}  & \xmark & \cmark & \xmark & Image, Text & Vision-Language applications & \xmark & CLIP-based \\

Zhao et al.~\cite{zhao2023identification}  &  \xmark & \cmark & \xmark & RGB & Disease Classification & \cmark & AgriCL \\

\textbf{MCLC (Ours)}  &  \cmark & \cmark & \cmark & \textbf{RGB, Depth, Thermal} & Leaf Count & \cmark & \textbf{MMLC (our dataset)} \\
\bottomrule
\end{tabular}}
\end{table*}

\section{Related Work} \label{sec:literature}

Leaf counting is important for plant phenotyping~\cite{buzzy2020real} as it helps monitor plant growth~\cite{shreyaTAFE}\cite{bonoTAFE}, development~\cite{donapati2023real_TAFE}, and health~\cite{gargTAFE}. Many deep learning-based methods have been proposed for automatic leaf-counting ~\cite{tu2020automatic}~\cite{tu2022toward}. In this section, we briefly review existing leaf-counting methods, the use of multimodal data for counting, and recent advances in CL, particularly in multimodal settings.

\textbf{Leaf counting methods.} \label{sec:literature_l1}
Existing leaf counting paradigms fall into four main categories: direct regression~\cite{giuffrida2016learning, giuffrida2018pheno, itzhaky2018leaf}, segmentation-based~\cite{Kuznichov_2019_CVPR_Workshops, deb2024cnn}, density estimation~\cite{aich2018improving, xie2018microscopy}, and object detection methods~\cite{tu2020automatic, lu2021counting}. While most works rely on RGB images, some have explored multimodal data (e.g., depth, thermal, NIR, fluorescence) to improve robustness~\cite{dobrescu2017leveraging, giuffrida2018pheno}. However, existing studies use static settings and do not address the challenges of learning from sequential data. In contrast, in this work, we explore leaf counting from a CL perspective using multimodal inputs, addressing both distribution shifts and scalability in real-world agricultural environments for the first time.

\textbf{Multimodal Continual Learning.} \label{sec:literature_l2}
CL methods are broadly categorized into regularization-based~\cite{kirkpatrick2017overcoming}, architectural-based~\cite{yoon2017lifelong}, and rehearsal-based approaches~\cite{lopez2017gradient,chaudhry2018efficient}. Regularization-based methods aim to preserve previously acquired knowledge by constraining updates to important model parameters, without requiring access to past data. Architectural-based methods allocate distinct portions of the network to different tasks, using either fixed-capacity or dynamically expanding architectures to accommodate new information. Rehearsal-based approaches mitigate forgetting by storing and replaying representative samples from prior tasks. Within this category, experience replay (ER)~\cite{rolnick2019experience} utilizes a memory buffer to retain raw data samples, while generative replay~\cite{wang2019continual} relies on generative models to synthesize past data. Although generative methods reduce storage requirements, they introduce significant complexity. In contrast, ER has proven to be simpler, more effective, and widely adopted in recent studies~\cite{bidaki2025online}, and is therefore explored in this work. A key challenge in ER is to select representative samples for the memory buffer. Earlier works, such as iCaRL \cite{rebuffi2017icarl}, use a herding strategy to select samples closest to the class means, approximating the data distribution with limited memory. Similarly, prototype-based methods maintain representative feature distributions for replay and knowledge retention \cite{zhang2019variational}. Also, Prototype-Guided Memory Replay \cite{ho2023prototype} further improves replay efficiency by using class-level prototypes to select or generate informative samples. However, these approaches are primarily designed for classification tasks, where clear class boundaries exist, and class prototypes can be defined. In contrast, our work focuses on a regression setting for leaf counting, where outputs are continuous and defining class prototypes is not straightforward. Therefore, instead of relying on class-wise representatives, we propose a hybrid uncertainty-diversity sampling strategy, which selects samples based on prediction uncertainty and feature-space diversity. This enables the model to retain both informative and non-redundant samples, making it more suitable for multimodal regression-based CL.

Some recent efforts have extended multimodal CL to other domains. Sun et al. update features and knowledge across tasks, even when modalities are missing~\cite {sun2020multimodal}. Wang et al. focus on cross-modal retrieval with continual indexing~\cite{wang2021continual}, while a general framework has been developed for vision-language tasks~\cite{srinivasan2022climb}. Other studies have explored multimodal CL in robotics \cite{lesort2020continual}, activity monitoring~\cite{gai2021multi, cai2024dynamic}, and agriculture \cite{li2023labelCOMPAG}. 
In agriculture~\cite{dimitriTAFE,li2021metaCOMPAG}, CL has been applied to plant disease detection~\cite{zhao2023identification, page2023class,li2024rehearsalCOMPAG}, plant recognition~\cite{chien2024plant}, and stress classification~\cite{reza2024optimizing}, but to the best of our knowledge, CL in leaf counting is still underexplored. Although CL is widely explored in classification, its use in multimodal regression remains underexplored~\cite{he2021clear}. This work presents the first CL approach for leaf counting, using multimodal data to support continual adaptation in dynamic agricultural environments.

From Table \ref{tab:comparison_lit_final}, we observe that earlier methods mainly used RGB images and focused on static datasets for tasks such as leaf counting or segmentation. Although some included additional modalities such as NIR or fluorescence, none addressed CL, which provides temporal data across tasks. Some recent works have explored CL, but mostly for classification tasks using single-modality data or non-visual inputs. These are often not designed for plant counting and rarely use real field data. To the best of our knowledge, no prior work has explored leaf counting in a CL setting using multimodal inputs, making our approach a novel contribution that addresses an existing gap in the literature. Our work uses RGB, depth, and thermal modalities for leaf-counting. This addresses both the need for multimodal fusion and continual adaptation in a real-world agricultural setting, enabling sequential learning over time.

\section{Proposed Dataset} \label{sec:dataset}

We introduce a novel MMLC dataset for leaf-counting under a DIS in CL. The dataset contains $6.3K +$ images across RGB, depth, and thermal modalities, captured in real agricultural fields at different times of day. The datasets include four vegetable crop types, including \textit{capsicum}, \textit{zucchini}, \textit{cucumber}, and \textit{cauliflower}, showing diversity in shape and appearance. The samples were collected within the first four weeks of plant growth, a crucial period for early phenotypic analysis. The image triplets, consisting of RGB, depth, and thermal images, were captured with a \textbf{Google Pixel 5 smartphone camera}, an \textbf{Intel® RealSense™ D435 sensor}, and a \textbf{Fluke TiX580 infrared sensor}, respectively. All plant images were acquired from a consistent top-down perspective at approximately 1 meter to ensure uniformity in data collection. To incorporate natural environmental variation, the plants were captured twice a day, once in the morning and once in the evening, enabling the dataset to reflect changes in illumination and temperature throughout the day.

\begin{figure*}[!t]
\centering
\includegraphics[width=0.97\linewidth]{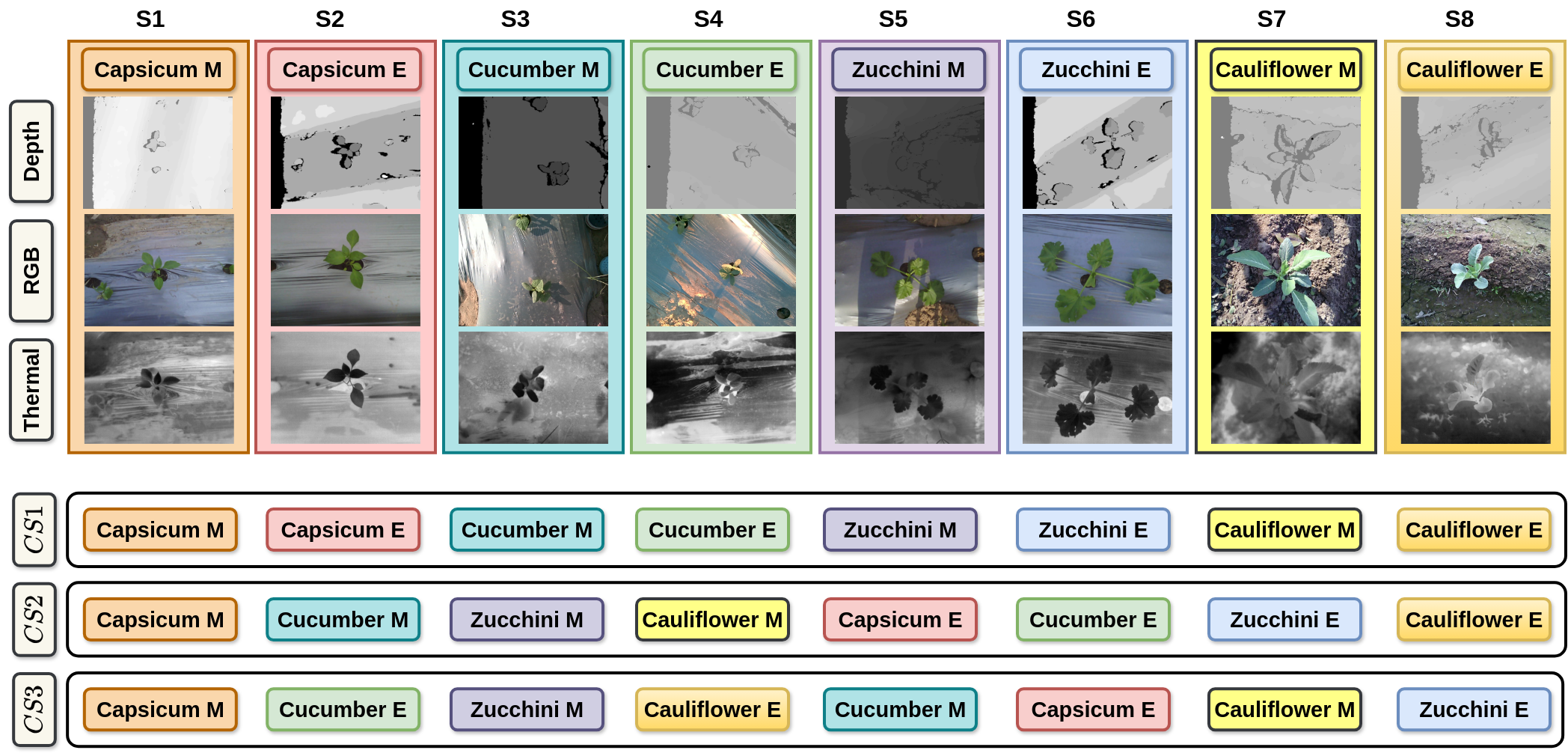}
\caption{
Overview of the proposed MMLC dataset. \textit{Top}: It shows example images (RGB, depth, and thermal) from the 8 splits created from the MMLC dataset. \textit{Bottom}: It presents the order of the splits in our 4 crops created for a CL setting. The three crop sequences $CS1$, $CS2$, and $CS3$ are formed by following an ordering by crop type, capture time, and mixed, respectively.  }
\label{fig:tasks}
\end{figure*}

To enable CL benchmarking and evaluation on our MMLC dataset, we adopt a DIS in which tasks within a given sequence exhibit potential data distribution shifts, also termed "domain shifts" \cite{kumari2025continual}. In the MMLC dataset, the variation in data arises either due to changes in the crop species (\textit{capsicum}, \textit{zucchini}, \textit{cucumber}, and \textit{cauliflower}) or the lighting/thermal conditions based on the time of capture (\textit{morning or evening}). We propose a partition of the MMLC dataset into eight splits (S1, $\dots$, S8), treated as eight tasks. Specifically, we group samples by crop type and capture time. Thus, we obtain a total of 8 tasks across 4 crop types and 2 capture times. A few samples from each of the splits are shown in Figure~\ref{fig:tasks}.  Other splits, e.g., grouping samples only by crop type or capture time, can be used to simulate different DIS.

For an exhaustive CL benchmarking, we provide three possible orderings of the eight splits, termed $CS1$, $CS2$, and $CS3$, from MMCL as discussed below: 

\noindent \textbf{$CS1$ (Crop-wise):} In this crop sequence, tasks are organized based on crop type. The first two tasks include images of capsicum for the morning and evening, respectively. Tasks 3 and 4 contain morning and evening images of a cucumber; tasks 5 and 6 contain morning and evening images of a zucchini; and tasks 7 and 8 contain morning and evening images of a cauliflower. This setup ensures the model is exposed to one crop at a time across different times of the day before moving to the next crop.

\noindent  \textbf{$CS2$ (Time-wise):} Here, tasks are arranged based on the time of image capture. Tasks 1 to 4 consist of morning images of capsicum, cucumber, zucchini, and cauliflower, respectively. Tasks 5 to 8, then present the evening images of these same crops in the same order. This setup allows the model to first learn representations influenced by time of day, then generalize across crops.

\noindent \textbf{$CS3$ (Mixed):} This sequence introduces a mix of crop types and capture times in a non-linear order. Task 1 contains morning images of capsicum; task 2 has evening images of cucumber; task 3 introduces a new crop, zucchini, in the morning; and task 4 has another new crop, cauliflower, in the evening. Tasks 5 to 8 revisit the previous crops at different times: morning images of cucumber, evening images of capsicum, morning images of cauliflower, and evening images of zucchini. This sequence mimics a more realistic and unpredictable data stream.

\section{Methodology} \label{sec:method}
\begin{figure*}[htbp]
\centering
\includegraphics[width=1\linewidth]{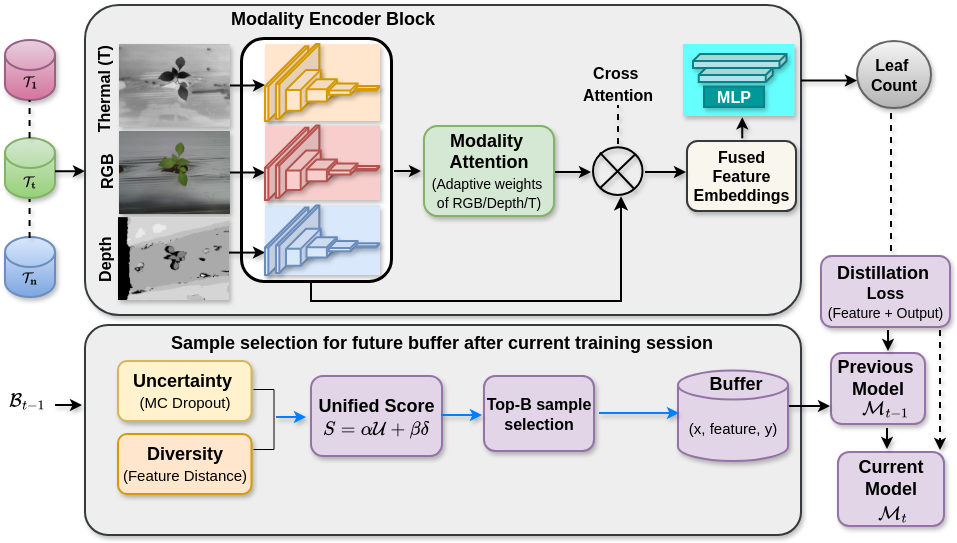}
\caption{Overview of \textbf{Multimodal Continual Leaf Counting (MCLC)} framework. 
It processes paired RGB, depth, and thermal images to predict leaf count. Features from each modality are extracted using a modality encoder block, adaptively weighted via modality attention, fused using cross-attention, and passed to an MLP regressor for prediction. To mitigate catastrophic forgetting, the model is trained on the current task $T_t$ and on memory samples from a buffer $B_{t-1}$ that stores input samples, feature representations, and labels. After each training session, samples from $T_t$ are evaluated using a unified uncertainty-diversity score, where uncertainty is estimated via Monte Carlo dropout and diversity via feature embedding distance. The top-$B$ samples are stored in the buffer for future training. Additionally, knowledge from previous tasks is preserved using distillation between $M_{t-1}$ and $M_t$.}

\label{fig:framework}
\end{figure*}

We address the challenge of leaf counting in real-world agricultural environments, where non-stationary data distributions arise from changing environmental and plant conditions. To ensure robust and consistent performance in such dynamic scenarios, we propose an MCLC framework illustrated in Figure \ref{fig:framework} that combines multimodal feature fusion with a CL strategy. In the following subsections, we formally define the problem, describe the proposed MCLC pipeline, and introduce the CL mechanism designed to mitigate catastrophic forgetting. The complete training procedure is summarized in Algorithm~\ref{algo:mcc}.

\begin{algorithm}[!t]
\caption{Multimodal Continual Learning}
\label{algo:mcc}
\renewcommand{\arraystretch}{1.2}
\begin{algorithmic}[1]
\STATE \textbf{Input:} Tasks $\{\mathcal{T}_1, \dots, \mathcal{T}_n\}$ with data volume $\{\mathcal{D}_1, \dots, \mathcal{D}_n\}$; buffer size $B$
\STATE \textbf{Output:} Final model $\mathcal{M}_n$ trained with replay

\STATE Initialize memory buffer $\mathcal{B}_0 \gets \emptyset$
\STATE Initialize model $\mathcal{M}_0$

\FOR{$t = 1$ to $n$}

    \STATE \textbf{Step 1: Train model on current task with replay and distillation}
    \FOR{each mini-batch $(x, y) \sim \mathcal{D}_t$}
        \STATE Compute prediction $\hat{y} \gets \mathcal{M}_t(x)$
        \STATE Compute task loss $\ell_t \gets \ell(\hat{y}, y)$

        \STATE \textbf{Replay:} Sample $(x_b, f_b, y_b)$ from $\mathcal{B}_{t-1}$
        \STATE $\ell_{\text{replay}} \gets \ell(\mathcal{M}_t(x_b), y_b) + \ell(\mathcal{R}(f_b), y_b)$

        \IF{$t > 1$}
            \STATE Compute distillation loss:
            \STATE $\ell_{\text{distill}} \gets \|\phi_t(x) - \phi_{t-1}(x)\|_2^2 + \|\mathcal{M}_t(x) - \mathcal{M}_{t-1}(x)\|_2^2$
        \ENDIF

        \STATE Update model using:
        \STATE $\ell \gets \ell_t + \lambda_1 \ell_{\text{replay}} + \lambda_2 \ell_{\text{distill}}$
    \ENDFOR

    \STATE \textbf{Step 2: Update memory buffer}
    \FOR{each $(x_t^i, y_t^i) \in \mathcal{D}_t$}
        \STATE Compute uncertainty via MC Dropout:
        \STATE $\mathcal{U}(x_t^i) \gets \text{Var}(\{\mathcal{M}_t^{(k)}(x_t^i)\}_{k=1}^{T})$
        \STATE Compute feature embedding $f_t^i \gets \phi(x_t^i)$

        \STATE Compute diversity:
        \STATE $\delta(x_t^i) \gets \min_{x' \in \mathcal{B}_{t-1}} \| f_t^i - \phi(x') \|_2$

        \STATE Compute score:
        \STATE $S(x_t^i) \gets \alpha \mathcal{U}(x_t^i) + \beta \delta(x_t^i)$
    \ENDFOR

    \STATE Select top-$B$ samples based on $S(x)$ to form $\mathcal{B}_t$
    \STATE Store $(x, f, y)$ in $\mathcal{B}_t$

    \STATE Freeze current model: $\mathcal{M}_{t-1} \gets \mathcal{M}_t$

\ENDFOR

\STATE \textbf{Return:} Final model $\mathcal{M}_n$
\end{algorithmic}
\end{algorithm}

\subsection{Problem Formulation} \label{subsec:problem}

We consider a DIS in which a model is sequentially trained on different tasks (or episodes), each of which may involve domain shifts. Let there be $n$ tasks denoted as $\{\mathcal{T}_1, \mathcal{T}_2, \dots, \mathcal{T}_n\}$, where each task $\mathcal{T}_t$  corresponds to the $t^{th}$ domain, i.e., a combination of a plant species and imaging conditions in our setting. Each task is associated with a data volume $\mathcal{D}_t = \{(x^i_t, y^i_t)\}_{i=1}^{N_t},$ where $N_t$ is the total number of samples in $\mathcal{D}_t$, $x_t^i$ represents a multimodal input and $y_t^i$ is the corresponding leaf count.
Each input $x_t^i$ comprises three modalities: RGB, thermal, and depth. Specifically, we denote
\[    x_t^i = \left\{ x_t^{i, m} \mid m \in \{\text{RGB},\ \text{Thermal},\ \text{Depth}\}\right\} \in \mathcal{X}_t, 
\]
where $x_t^{i, \text{RGB}} \in \mathbb{R}^{H \times W \times 3}$ is a 3-channel image, and $x_t^{i, \text{Thermal}},\ x_t^{i, \text{Depth}} \in \mathbb{R}^{H \times W}$ are single-channel images. Here, $H$ and $W$ denote the image height and width, respectively. Each ground truth label $y_t^i \in \mathbb{R}$. For simplicity, we will refer to the multimodal input as $x_t^i$ throughout the rest of the paper, with the understanding that it implicitly includes all three modalities: RGB, thermal, and depth, unless explicitly stated otherwise.

Ideally, the goal is to learn a model $\mathcal{M}_\theta$, parameterized by $\theta$, that minimizes the cumulative loss across all tasks in the sequence ($\{\mathcal{T}_1, \mathcal{T}_2, \dots,\mathcal{T}_n\}$) for leaf counting (Eq.~\ref{eq:joint_eq}).
\begin{equation}
\label{eq:joint_eq}
\arg\min_{\theta} \sum_{t=1}^{n} \sum_{(x^i_t, y^i_t) \in \mathcal{D}_t} \ell(\mathcal{M}_\theta(x^i_t), y^i_t)
\end{equation}
where $\ell$ denotes the regression loss function, which we define as Mean Squared Error (MSE) in our setting.

This objective cannot be directly optimized, as the model has access only to the current data volume $\mathcal{D}_t$ at the $t^{th}$ training session. So, our objective is to model each task $\mathcal{T}_t$ in sequence using a deep regression model $\mathcal{M}_t$ with parameters $\theta_t$, such that performance on previous tasks is preserved. Since previous task data is not retained, we store a subset of samples in a buffer $\mathcal{B}_t$ during training. This buffer enables each new model $\mathcal{M}_{t+1}$ to retain knowledge of prior tasks $\{\mathcal{T}_1, \dots, \mathcal{T}_t\}$. This buffer $\mathcal{B}_t$ is constructed by sampling from the current data volume $\mathcal{D}_t$  based on a hybrid strategy combining \textit{uncertainty} and \textit{diversity} (see Sec.~\ref{subsec:CL}). Specifically, we compute uncertainty using multiple stochastic forward passes with dropout activated, which captures prediction variability. Further, instead of storing only raw samples, we store both input samples and their corresponding fused feature representations. The selected samples are then added to the buffer and used in future training to mitigate forgetting.

At each task $T_t$, we train the model $\mathcal{M}_t$ on buffer-aware data volume $\widetilde{\mathcal{D}}_t$,  the union of the current data volume and buffered samples from previous tasks as given in (Eq.~\ref{eq:unified D}).
\begin{equation}
   \widetilde{\mathcal{D}}_t = \mathcal{D}_t \cup  \mathcal{B}_{t-1}
   \label{eq:unified D}
\end{equation}

The learning objective is to minimize the MSE for each task $T_t$ (Eq.~\ref{equ: loss})
\begin{equation}
    \ell_t = \frac{1}{|\widetilde{\mathcal{D}}_t|} \sum_{(x, y) \in \widetilde{\mathcal{D}}_t} \left( \mathcal{M}_t(x; \theta_t) - y \right)^2
    \label{equ: loss}
\end{equation}

where $\ell_t$ denotes the regression loss for task $T_t$, implemented using Smooth L1 loss for improved stability against outliers. While only the current data volume $\mathcal{D}_t$ and buffered samples from $\mathcal{B}_{t-1}$ are used for training at each step, we evaluate the final model $\mathcal{M}_n$ on all task data volumes $\{\mathcal{D}_1, \dots, \mathcal{D}_n\}$ to assess overall continual performance. 

\subsection{Multimodal Leaf Counting} \label{sec:mmlc}
Our MCLC framework comprises four main components: a modality-encoder block (MEB), a modality attention block, a cross-attention fusion module, and a regression head. The architecture processes multimodal inputs to predict a single leaf count, as summarized in Algorithm~\ref{algo:mclc}.


\textbf{Modality Encoder Block:}  
Each modality, RGB, depth, and thermal, is processed by a dedicated modality encoder (ME) based on the ResNet-50 architecture. Let the input image for modality $m \in \{\text{RGB}, \text{Depth}, \text{Thermal}\}$ be denoted as $x_t^{i, m}$.

We define the modality encoder $f_m(\cdot)$ and a projection head $g_m(\cdot)$ for each modality such that:
\begin{equation}
z_t^{i, m} = g_m\left(f_m\left(x_t^{i, m}\right)\right), \quad z_t^{i, m} \in \mathbb{R}^{512}
\end{equation}

where:
$f_m(\cdot)$ is the modality-specific encoder (ResNet-50-based), and $g_m(\cdot)$ is a projection head comprising a flattening layer and a fully connected layer that projects the 2048-dimensional feature to a 512-dimensional embedding, with ReLU activation and dropout regularization. $z_t^{i, m}$ is the resulting 512-dimensional embedding for the modality $m$ of the sample $i$ from the task $\mathcal{T}_t$.


 

\textbf{Modality Attention:}
Before fusion, we introduce a modality attention mechanism to adaptively weight the contribution of each modality. Given embeddings $z_t^{i,\text{RGB}}, z_t^{i,\text{Depth}}, z_t^{i,\text{Thermal}}$, we compute attention weights as:
\begin{equation}
\begin{aligned}
[w_{\text{RGB}}, w_{\text{Depth}}, w_{\text{Thermal}}] 
= \text{Softmax}\Big( 
h([z_t^{i,\text{RGB}}, \\
z_t^{i,\text{Depth}}, z_t^{i,\text{Thermal}}]) 
\Big)
\end{aligned}
\end{equation}


There $h(\cdot)$ is a two-layer MLP followed by Softmax to generate normalized attention weights for the RGB, depth, and thermal modalities. The re-weighted embeddings are:
\begin{equation}
\tilde{z}_t^{i,m} = w_m \cdot z_t^{i,m}
\end{equation}

This allows the model to dynamically focus on more informative modalities.

\textbf{Cross-Attention Fusion:}  
Different modality embeddings are fused using a multi-head cross-attention mechanism. The RGB embedding serves as the query (Q), while the depth and thermal embeddings act as keys (K) and values (V). We stack the depth and thermal embeddings to form the K and V matrices:
\begin{equation}
   K = V = \text{Stack}\left(z_t^{i, \text{Depth}},\ z_t^{i, \text{Thermal}}\right) \in \mathbb{R}^{2 \times 512} 
\end{equation}

The cross-attention module (CA) computes the fused representation as:
\begin{equation}
\phi(x_t^i) = {z}_t^i = \text{CA}\left(Q = z_t^{i, \text{RGB}},\ K,\ V\right) \in \mathbb{R}^{1 \times 512}
\end{equation}

where \text{CA} denotes a multi-head cross-attention mechanism with 4 attention heads operating on 512-dimensional modality embeddings and $\phi(\cdot)$ denotes the fused feature extractor for input $x_t^i$. The output ${z}_t^i$ is a fused 512-dimensional feature vector containing integrated multimodal information.

\textbf{Regression Head}: The fused features  ${z}_t^i$ are passed through a multi-layer perceptron to predict the leaf count as a single scalar value $\hat {y} _ t^i$.

\begin{algorithm}[!t]
\caption{Multimodal Leaf Counting $\mathcal{M}_t$ }

\label{algo:mclc}
\begin{algorithmic}[1]
\STATE \textbf{Input:} Multimodal sample \\ $x_t^i = \{x_t^{i,\text{RGB}}, x_t^{i,\text{Depth}}, x_t^{i,\text{Thermal}}\} \in \widetilde{\mathcal{D}}_t$

\STATE \textbf{Step 1: Modality Encoding}
\FOR{each modality $m \in \{\text{RGB}, \text{Depth}, \text{Thermal}\}$}
    \STATE $z_t^{i,m} \gets g_m(f_m(x_t^{i,m})) \in \mathbb{R}^{512}$
\ENDFOR

\STATE \textbf{Step 2: Modality Attention}
\STATE $[w_{\text{RGB}}, w_{\text{Depth}}, w_{\text{Thermal}}] \gets \text{Softmax}(h([z_t^{i,\text{RGB}}, 
   z_t^{i,\text{Depth}}, z_t^{i,\text{Thermal}}]))$
\STATE $\tilde{z}_t^{i,m} \gets w_m \cdot z_t^{i,m}$

\STATE \textbf{Step 3: Cross-Attention Fusion}
\STATE $Q \gets \tilde{z}_t^{i,\text{RGB}}$
\STATE $K \gets \text{Stack}(\tilde{z}_t^{i,\text{Depth}},\ \tilde{z}_t^{i,\text{Thermal}})$
\STATE $V \gets K$
\STATE $z_t^i \gets \text{CA}(Q, K, V)$

\STATE \textbf{Step 4: Regression}
\STATE $\hat{y}_t^i \gets \text{RegHead}(z_t^i)$

\STATE \textbf{Output:} Predicted leaf count $\hat{y}_t^i$
\end{algorithmic}
\end{algorithm}

\subsection{Continual Learning Strategy} \label{subsec:CL}
To mitigate catastrophic forgetting, we devise a novel CL strategy in our MCLC framework. The key components are:

 \textbf{Experience Replay (ER)}: To mimic the past tasks, a global buffer is maintained with a fixed allowed size, $\mathcal{B}$. The model is trained not only with the new data volume $\mathcal{D}_t$ but also with the buffer $\mathcal{B}_{t-1}$  which contains a few exemplars from already seen tasks $\{\mathcal{T}_1, \dots, \mathcal{T}_t\}$. In each training batch, we randomly sample a subset of stored samples from $\mathcal{B}_{t-1}$. The replay mechanism operates at two levels:
\textit{(1) Feature-level replay}, where stored fused features are directly used to train the regression head, and
\textit{(2) Input-level replay}, where stored raw samples are passed through the full model.

The combined replay loss is defined as:
\begin{equation}
\ell_{\text{replay}} = \lambda_1 \cdot \ell(\mathcal{R}(f), y) + \lambda_2 \cdot \ell(\mathcal{M}_t(x), y)
\end{equation}

where $f$ denotes stored fused features, $\mathcal{R}$ is the regression head, and $\lambda_1, \lambda_2$ are weighting factors. In our implementation, the weighting factors are set to $\lambda_1=0.3$ and $\lambda_2=0.7$ for feature-level and input-level replay, respectively.

 \textbf{Uncertainty-Diversity Sampling}: When adding new samples to the buffer, we propose to prioritize those with high uncertainty and diversity. We measure uncertainty ($\mathcal{U}(x_t^i)$) using Monte Carlo dropout, which captures prediction variability under stochastic forward passes. Specifically, we perform $T$ forward passes with dropout enabled and compute the variance of predictions:
\begin{equation}
\mathcal{U}(x_t^i) = \text{Var} \left( \{ \mathcal{M}_t^{(k)}(x_t^i) \}_{k=1}^{T} \right)
\end{equation}

where $\mathcal{M}_t^{(k)}$ denotes the model with dropout active during the $k^{th}$ forward pass. In our implementation, uncertainty is estimated using 5 stochastic forward passes with dropout enabled during inference.

Further, instead of a strict two-stage filtering, we compute a unified score that jointly considers uncertainty and diversity. For each candidate sample $(x_t^i, y_t^i)$, we compute a diversity score using Minimum Embedding Distance (MED) with respect to the existing buffer:
\begin{equation}
\delta(x_t^i) = \min_{x' \in \mathcal{B}_{t-1}} \left\| \phi(x_t^i) - \phi(x') \right\|_2
\end{equation}

The final selection score is computed as:
\begin{equation}
S(x_t^i) = \alpha \cdot \mathcal{U}(x_t^i) + \beta \cdot \delta(x_t^i)
\end{equation}

where $\alpha$ and $\beta$ are weighting factors. The top $\mathcal{B}$ samples with the highest scores are selected to update the buffer:
\begin{equation}
\mathcal{B}_t = \underset{x \in \widetilde{\mathcal{D}}_t}{\arg\max_B} \ S(x)
\end{equation}

The buffer maintains a fixed memory and is updated after each task by retaining samples with the highest combined uncertainty-diversity scores.
This unified score ensures that selected samples are both uncertain (informative) and diverse (less redundant w.r.t.\ past memory), thereby improving the quality of the buffer in CL.


\textbf{Knowledge Distillation:}  
To further preserve knowledge from previous tasks, we incorporate a distillation mechanism. After each task, a frozen copy of the previous model $\mathcal{M}_{t-1}$ is maintained. During training of $\mathcal{M}_t$, we enforce consistency between the current and previous model in both feature and output space:
\begin{equation}
\ell_{\text{distill}} = \left\| \phi_t(x) - \phi_{t-1}(x) \right\|_2^2 + \left\| \mathcal{M}_t(x) - \mathcal{M}_{t-1}(x) \right\|_2^2
\end{equation}

This helps stabilize representations and reduce catastrophic forgetting.

The overall training objective combines current task loss, replay loss, and distillation loss:
\begin{equation}
\ell = \ell_t + \ell_{\text{replay}} + \ell_{\text{distill}}
\end{equation}

\section{ Experiments and results} \label{sec:experiment_results}
In this section, we describe the experimental setup, the compared methods,  the evaluation metrics, and the computational complexity. We then provide a quantitative and qualitative analysis of the proposed model's performance relative to baseline methods and state-of-the-art approaches.

\subsection{Experimental Setup} \label{subsec:exp}
We evaluate our proposed method, MCLC, in a CL setting for multimodal leaf counting. Each task contains images from all three modalities, with the objective of predicting the leaf count.

\textbf{Training and implementation details.} 
All methods are trained sequentially across tasks using Smooth L1 and Mean Squared Error (MSE) losses for leaf counting. The buffer is used to retain the representative samples selected using the hybrid uncertainty-diversity criterion. To ensure fairness and a statistically reliable comparison, all experiments are repeated using three random seeds (42, 123, and 999). The reported results correspond to the mean and standard deviation across these three runs. The additional implementation details of the MCLC are given in Table \ref{tab:implementation}.

\begin{table}[!ht]
\centering
\caption{Implementation details}
\label{tab:implementation}
\renewcommand{\arraystretch}{1.2}
\begin{tabular}{lc}
\toprule
\textbf{Parameter} & \textbf{Value} \\
\midrule
Backbone Network & ResNet-50 \\
Embedding Dimension & 512 \\
Attention Heads & 4 \\
Optimizer & Adam \\
Learning Rate & $10^{-4}$ \\
Batch Size & 8 \\
Training Epochs & 30 \\
Dropout Rate & 0.3 \\
Buffer Size & \{30,40,50,80\} \\
MC Dropout Passes & 5 \\
Replay Weights & \{0.3, 0.7\} \\
Distillation Weight & 0.05 \\
Random Seed & \{42, 123, 999\} \\
GPU & NVIDIA H100 \\
\bottomrule
\end{tabular}
\end{table}

\textbf{Datasets Used}.
We evaluate our approach and existing works using three datasets: MMLC, MSU-PID, and CVPPP. The MMLC dataset is our proposed benchmark for multimodal CL in leaf counting and has been described in detail in section \ref{sec:dataset}. It comprises RGB, depth, and thermal modalities. For experimental evaluation, we consider three dataset sequences, CS1, CS2, and CS3, to simulate different domain incremental CL settings. We consider the MSU-PID dataset~\cite{cruz2016multi}, a publicly available multimodal leaf-counting dataset comprising RGB, thermal, and fluorescence images. Since it was not originally designed for CL evaluation, we adapted it to the CL setting by creating six sequential splits. MSU-PID contains approximately 1,000 samples from two crops (Arabidopsis and Beans), captured hourly over a period of 16 days. It serves as an additional benchmark for evaluating multimodal leaf-counting performance. Similar to MMLC, we adopt an 80:20 train:test division for each split. Additionally, we include the unimodal publicly available CVPPP dataset~\cite{bell2016aberystwyth}, a widely used benchmark for leaf counting based on RGB images of Arabidopsis and Tobacco plants. We used CVPPP to evaluate different leaf-counting paradigms. The detailed statistics and task distributions of the multimodal datasets are reported in Table~\ref{tab:task_stats}.

\begin{table}[!hpb]
    \centering
    \renewcommand{\arraystretch}{2.2} 
    \caption{Train-test details of the splits curated from the MSU-PID~\cite{cruz2016multi} and the MMLC datasets for CL experiments.}
    \label{tab:task_stats}
    \scalebox{0.85}{
    \begin{tabular}{c|c|c|c|c|c|c|c|c|c}
        \hline
        & \textbf{Splits}& \textbf{S1} & \textbf{S2} & \textbf{S3} & \textbf{S4} & \textbf{S5} & \textbf{S6} & \textbf{S7} & \textbf{S8} \\\hline
        \multirow{2}{*}{\rotatebox{90}{\shortstack{\textbf{MMLC}}}} 
        
        & Train  & 533 & 664  & 637  & 609  & 617 & 663 & 590 & 758  \\ 
        & Test & 133 & 166 & 158 & 151 & 153 & 165 & 146 & 187 \\  \hline
        \multirow{2}{*}{\rotatebox{90}{\shortstack{ \textbf{MSU-PID}}}} 
        & Train & 87 & 88 & 88 & 88 & 88 & 85 & - & - \\ 
        & Test  & 38 & 38 & 38 & 38 & 38 & 37 & - & -  \\ \hline
    \end{tabular}
    }    
\end{table}
 
\subsection{Comparable Methods} \label{baselines}
We compare two types of baselines, rehearsal-based and regularization-based methods, and a recent state-of-the-art approach. Rehearsal-based methods include ER \cite{rolnick2019experience}, which stores past examples, and DER++ \cite{buzzega2020dark}, which extends ER by enforcing consistency in predictions across time. Both are evaluated with memory sizes of \{${30, 40, 50, 80}$\}. The GEM \cite{lopez2017gradient} and A-GEM \cite{chaudhry2018efficient} control interference by projecting gradients using stored samples per task. We experiment with buffer sizes \{${4, 5, 6, 10}$\} per task for GEM and A-GEM. Regularization-based methods include EWC \cite{kirkpatrick2017overcoming}, which constrains parameter updates to those important for previous tasks. These are evaluated with regularization coefficients $\lambda \in \{{1, 2, 3}$\}. We also include the recent state-of-the-art method AVQACL \cite{wu2025avqacl}, which performs CL through adaptive quantization and contrastive learning and is evaluated with buffer sizes of \{${30, 40, 50, 80}$\}. In addition, we report results for three non-continual baselines: a naive model trained sequentially without any forgetting mitigation, which is considered a lower bound; cumulative training, where a model is retrained on all data seen so far; and joint training on the union of all task data, both of which are considered an upper bound.

\begin{table}[H]
\caption{Train-test performance matrix for $n=6$.}
\label{tab:matrix}
\centering
\renewcommand{\arraystretch}{1.25}

\begin{tabular}{c|cccccc}
\toprule
\textbf{Train$\backslash$Test} & $\mathbf{Te_1}$ & $\mathbf{Te_2}$ & $\mathbf{Te_3}$ & $\mathbf{Te_4}$ & $\mathbf{Te_5}$ & $\mathbf{Te_6}$ \\
\hline
$\mathbf{Tr_1}$ & $R_{1,1}$ & $R_{1,2}$ & $R_{1,3}$ & $R_{1,4}$ & $R_{1,5}$ & $R_{1,6}$ \\
$\mathbf{Tr_2}$ & $R_{2,1}$ & $R_{2,2}$ & $R_{2,3}$ & $R_{2,4}$ & $R_{2,5}$ & $R_{2,6}$ \\
$\mathbf{Tr_3}$ & $R_{3,1}$ & $R_{3,2}$ & $R_{3,3}$ & $R_{3,4}$ & $R_{3,5}$ & $R_{3,6}$ \\
  $\mathbf{Tr_4}$ & $R_{4,1}$ & $R_{4,2}$ & $R_{4,3}$ & $R_{4,4}$ & $R_{4,5}$ & $R_{4,6}$ \\
$\mathbf{Tr_5}$ & $R_{5,1}$ & $R_{5,2}$ & $R_{5,3}$ & $R_{5,4}$ & $R_{5,5}$ & $R_{5,6}$ \\
$\mathbf{Tr_6}$ & $R_{6,1}$ & $R_{6,2}$ & $R_{6,3}$ & $R_{6,4}$ & $R_{6,5}$ & $R_{6,6}$ \\
\bottomrule
\end{tabular}
\end{table}

\subsection{Evaluation metrics}
After training on the $t^{\text{th}}$ task using training data $Tr_t$, we evaluate the model on all test data $Te_1, Te_2, \dots, Te_n$. This results in a train-test matrix $R \in \mathbb{R}^{n \times n}$, where $R_{t,j}$ denotes the MSE on the test task $j$ after training on tasks up to $t$. The sample train-test matrix is shown in the Table \ref{tab:matrix}. This matrix captures the stability-plasticity trade-off \cite{mermillod2013stability} between the model's ability to acquire new knowledge (plasticity) and its ability to retain prior knowledge (stability).

To evaluate performance, we used standard CL metrics commonly used in the literature. For overall performance, we compute the average mean squared error (AMSE) across all tasks after completing the final task $T$ based on \cite{lopez2017gradient}:
\begin{equation}
\text{AMSE} = \frac{1}{T} \sum_{j=1}^{T} R_{T,j}
\end{equation}

Further, Backward Transfer (BWT) and Forward Transfer (FWT) \cite{diaz2018don} are used to evaluate forgetting and knowledge transfer. The lower the value, the better the model's performance.

\subsection{Computational Complexity and Scalability}

The proposed MCLC framework enables efficient sequential learning without retraining on all previously seen data. Instead, a bounded memory buffer is used to preserve representative samples from earlier tasks, thereby reducing storage and retraining requirements during continual adaptation. Using a memory buffer size of 80, the framework required an average training time of 13.06 minutes per task with a peak GPU memory usage of 3.75 GB. During inference, the model required only 10.68 ms per sample using a single forward pass. In addition, the modality attention and cross-attention modules introduced relatively low computational overhead compared to the backbone feature extractors. These characteristics show that MCLC can efficiently adapt to new tasks while maintaining low training, memory, and inference overhead, making it suitable for practical multimodal continual learning scenarios. 
\begin{table*}[htbp!]
\caption{Performance comparison of CL and non-CL methods across three \textbf{MMLC} dataset sequences (CS1, CS2, and CS3). For each method, the best-performing hyperparameters (Param.) are selected based on the lowest AMSE. Best and second-best CL performances are \textbf{Bold} and \uline{Underlined} respectively.}
\centering
\label{tab:mclc_best_results}
\scriptsize              
\setlength{\tabcolsep}{2.5pt}   
\renewcommand{\arraystretch}{1.15}

\resizebox{\textwidth}{!}{

\begin{tabular}{l|l|cccc|cccc|cccc}
\hline
 & \textbf{Method} &
\multicolumn{4}{c|}{\textbf{CS1}} &
\multicolumn{4}{c|}{\textbf{CS2}} &
\multicolumn{4}{c}{\textbf{CS3}} \\

& &
\textbf{Param.} & \textbf{BWT$\downarrow$} & \textbf{FWT$\downarrow$} & \textbf{AMSE$\downarrow$} &
\textbf{Parameters} & \textbf{BWT$\downarrow$} & \textbf{FWT$\downarrow$} & \textbf{AMSE$\downarrow$} &
\textbf{Parameters} & \textbf{BWT$\downarrow$} & \textbf{FWT$\downarrow$} & \textbf{AMSE$\downarrow$} \\
\hline

\multirow{7}{*}{\rotatebox[origin=c]{90}{\textbf{CL}}}

& EWC
& $\lambda=1$ & 2.058$\pm$0.447 & 2.402$\pm$0.602 & 3.110$\pm$1.626
& $\lambda=3$ & 2.845$\pm$1.244 & 3.255$\pm$1.356 & 4.905$\pm$5.179
& $\lambda=2$ &  18.935$\pm$29.857 & 19.310$\pm$30.015 & 2.328$\pm$0.237 \\

& GEM
& $\mathcal{B}$=10x8 & 0.647$\pm$0.084 &1.130$\pm$0.080 &1.160$\pm$0.063
& $\mathcal{B}$=10x8 & 2.917$\pm$3.656 & 1.398$\pm$0.199 & 1.107$\pm$0.054
& $\mathcal{B}$=10x8 & 0.680$\pm$0.159 &1.095$\pm$0.068 &1.167$\pm$0.291 \\

& A-GEM
& $\mathcal{B}$=10x8 & 0.733$\pm$0.217 &1.111$\pm$0.029 &0.954$\pm$0.061
& $\mathcal{B}$=6x8 & 1.057$\pm$0.106 &1.409$\pm$0.103 &1.055$\pm$0.071
& $\mathcal{B}$=10x8 & 0.676$\pm$0.056 &1.076$\pm$0.050 &1.082$\pm$0.057 \\

& DER++
& $\mathcal{B}$=40 &  1.576$\pm$1.703 & 21.301$\pm$1.866 & 24.462$\pm$2.159
& $\mathcal{B}$=80 & 1.108$\pm$0.463 & 22.119$\pm$0.163 & 25.157$\pm$0.929
& $\mathcal{B}$=50 & 1.758$\pm$0.786 & 21.613$\pm$2.484 & 26.132$\pm$2.456 \\

& ER
& $\mathcal{B}$=80 & \uline{0.565$\pm$0.144} & \uline{0.826$\pm$0.056} & \uline{0.843$\pm$0.090}
& $\mathcal{B}$=80 & \uline{0.564$\pm$0.069} & \uline{0.790$\pm$0.031} & \uline{0.743$\pm$0.049}
& $\mathcal{B}$=80 & \uline{0.394$\pm$0.038} & \uline{0.624$\pm$0.061} & \uline{0.749$\pm$0.120} \\

& AVQACL
& $\mathcal{B}$=50 & 1.368$\pm$0.300 & 2.084$\pm$0.281 & 2.241$\pm$0.114
& $\mathcal{B}$=50 & 1.422$\pm$0.145 & 2.154$\pm$0.121 & 1.585$\pm$0.128
& $\mathcal{B}$=40 & 1.325$\pm$0.267 & 1.979$\pm$0.236 & 1.561$\pm$0.115 \\

& \textbf{MCLC}
& $\mathcal{B}$=80 &  \textbf{0.280$\pm$0.115} & \textbf{0.661$\pm$0.104} & \textbf{0.675$\pm$0.027}
& $\mathcal{B}$=80 & \textbf{0.145$\pm$0.116} & \textbf{0.644$\pm$0.037} & \textbf{0.542$\pm$0.069}
& $\mathcal{B}$=80 & \textbf{0.055$\pm$0.038} & \textbf{0.594$\pm$0.050} & \textbf{0.745$\pm$0.057} \\

\hline

\multirow{3}{*}{\rotatebox[origin=c]{90}{\textbf{Non-CL}}}

& Naive
& --  & 1.570$\pm$0.510 & 1.854$\pm$0.474 & 1.518$\pm$0.303 
& -- & 1.966$\pm$0.713 & 2.428$\pm$0.543 & 2.674$\pm$1.873
& -- & 1.722$\pm$0.397 & 2.085$\pm$0.419 & 2.222$\pm$0.361 \\

& Cumulative
& -- & 0.128$\pm$0.106 & 0.303$\pm$0.135 & 0.232$\pm$0.072
& -- & 0.271$\pm$0.254 & 0.398$\pm$0.227 & 0.219$\pm$0.066
& -- & 0.090$\pm$0.049 & 0.153$\pm$0.027 & 0.157$\pm$0.32 \\

& Joint
& -- & -- & -- & 0.186$\pm$0.025
& -- & -- & -- & 0.193$\pm$0.015
& -- & -- & -- & 0.177$\pm$0.019 \\

\hline
\end{tabular}
}
\end{table*}
\subsection{Quantitative result analysis} \label{subsec:result}

\begin{table*}[htbp!]
\caption{Performance comparison of CL and non-CL methods on \textbf{MSU-PID} dataset. For each method, the best-performing hyperparameters are selected based on the lowest AMSE. Best and second-best CL performances are \textbf{Bold} and \uline{Underlined}, respectively.}
\centering
\label{tab:msupid_best_results}
\renewcommand{\arraystretch}{1.3}
\scalebox{1}{
\begin{tabular}{l|l|cccc}
\hline
  & \textbf{Method} & 
\multicolumn{4}{c}{\textbf{MSU-PID}} \\ 
& &
\textbf{Parameters} & \textbf{BWT~$\downarrow$} & \textbf{FWT~$\downarrow$} & \textbf{AMSE~$\downarrow$} \\
\hline

\multirow{7}{*}{\rotatebox{90}{\textbf{CL}}}
& EWC \cite{kirkpatrick2017overcoming} &  $\lambda$=2 & 1.062 $\pm$ 0.575  & 1.203 $\pm$ 0.105 & 0.882 $\pm$ 0.074 \\
& GEM \cite{lopez2017gradient} &  $\mathcal{B}$=10x8 & 1.134 $\pm$ 0.356 & 1.103 $\pm$ 0.067 & 0.869 $\pm$ 0.052  \\
& A-GEM \cite{chaudhry2018efficient} &   $\mathcal{B}$=10x8 & \uline{1.073 $\pm$ 0.574} & 1.116 $\pm$ 0.098  & 0.813 $\pm$ 0.084  \\
& DER++ \cite{buzzega2020dark} &    $\mathcal{B}$=80 & 3.656 $\pm$ 14.338  & 36.909 $\pm$ 6.826  & 37.616 $\pm$ 6.954  \\
& ER \cite{rolnick2019experience} &  $\mathcal{B}$=80 & 1.144 $\pm$ 0.590 & \uline{1.031 $\pm$ 0.133} & \uline{0.750 $\pm$ 0.109} \\
& AVQACL \cite{wu2025avqacl} &  $\mathcal{B}$=80 & \textbf{0.095 $\pm$ 0.292}  & 1.183 $\pm$ 0.186  & 1.030 $\pm$ 0.170  \\
& MCLC (Ours) &  $\mathcal{B}$=80 & 2.244 $\pm$ 0.148 & \textbf{0.979 $\pm$ 0.104} & \textbf{0.664 $\pm$ 0.071}  \\

\hline

\multirow{3}{*}{\rotatebox{90}{\textbf{Non-CL}}}
& Naive & -- & 0.948 $\pm$ 0.255 & 1.334 $\pm$ 0.145 & 1.097 $\pm$ 0.320 \\
& Cumulative & -- & 0.998 $\pm$ 0.354 & 1.032 $\pm$ 0.048 & 0.753 $\pm$ 0.118 \\
& Joint & -- & -- & -- &  0.642 $\pm$ 0.109  \\

\hline
\end{tabular}
}
\end{table*}

We experimented with two multimodal datasets, namely MMLC and MSU-PID, and compared the MCLC model against baselines and state-of-the-art methods. Table \ref{tab:mclc_best_results} shows the performance of the methods on the best-performing hyperparameters, $\lambda$ and buffer size on MCLC. Among all CL methods, the proposed MCLC consistently achieves the best performance across all three sequences. Specifically, MCLC obtains the lowest average AMSE of 0.675$\pm$0.027, 0.542$\pm$0.069, and 0.745$\pm$0.057 for CS1, CS2, and CS3, respectively. It also achieves the lowest BWT values of 0.280$\pm$0.115, 0.145$\pm$0.116, and 0.055$\pm$0.038, along with the lowest FWT values of 0.661$\pm$0.104, 0.644$\pm$0.037, and 0.594$\pm$0.050 for the respective task sequences, indicating improved knowledge retention and forward transfer. Among the CL approaches compared, ER achieves the second-best performance in most cases, followed by A-GEM. In contrast, EWC, DER++, and AVQACL produce comparatively higher AMSE, BWT, and FWT values. Although rehearsal-based methods generally outperform regularization-based approaches such as EWC, DER++ performs poorly in our experiments. This is because it combines memory with logit distillation, which is designed for classification. When applied to regression-based leaf counting, distilling continuous outputs propagates prediction errors across tasks, resulting in error accumulation and reduced stability. 

Table \ref{tab:msupid_best_results} shows the performance of the methods on the best-performing hyperparameters, $\lambda$ and buffer size on MSU-PID. Among all CL methods, the MCLC achieves the best overall performance, with the lowest average AMSE of 0.664$\pm$0.071 and the lowest FWT of 0.979$\pm$0.104, indicating better generalization to new tasks and improved forward knowledge transfer. Although AVQACL achieves the lowest BWT (0.095$\pm$0.292), MCLC provides a better balance between knowledge retention and adaptation, resulting in the best overall continual learning performance. ER achieves the second-best performance with an AMSE of 0.750$\pm$0.109 and an FWT of 1.031$\pm$0.133, demonstrating the effectiveness of rehearsal-based learning. In comparison, EWC, GEM, and A-GEM show higher errors, while DER++ performs significantly worse, as on the MMLC dataset. Among the non-CL methods, the Joint achieves an AMSE of 0.642$\pm$0.109, serving as an upper bound because it has access to all training data. Compared with all CL methods, the proposed MCLC achieves the lowest AMSE in the CL setting, demonstrating its robustness and effectiveness on the MSU-PID dataset.



\begin{figure*}[t]
\centering

\begin{tabular}{ccc}

\includegraphics[width=0.65\columnwidth]{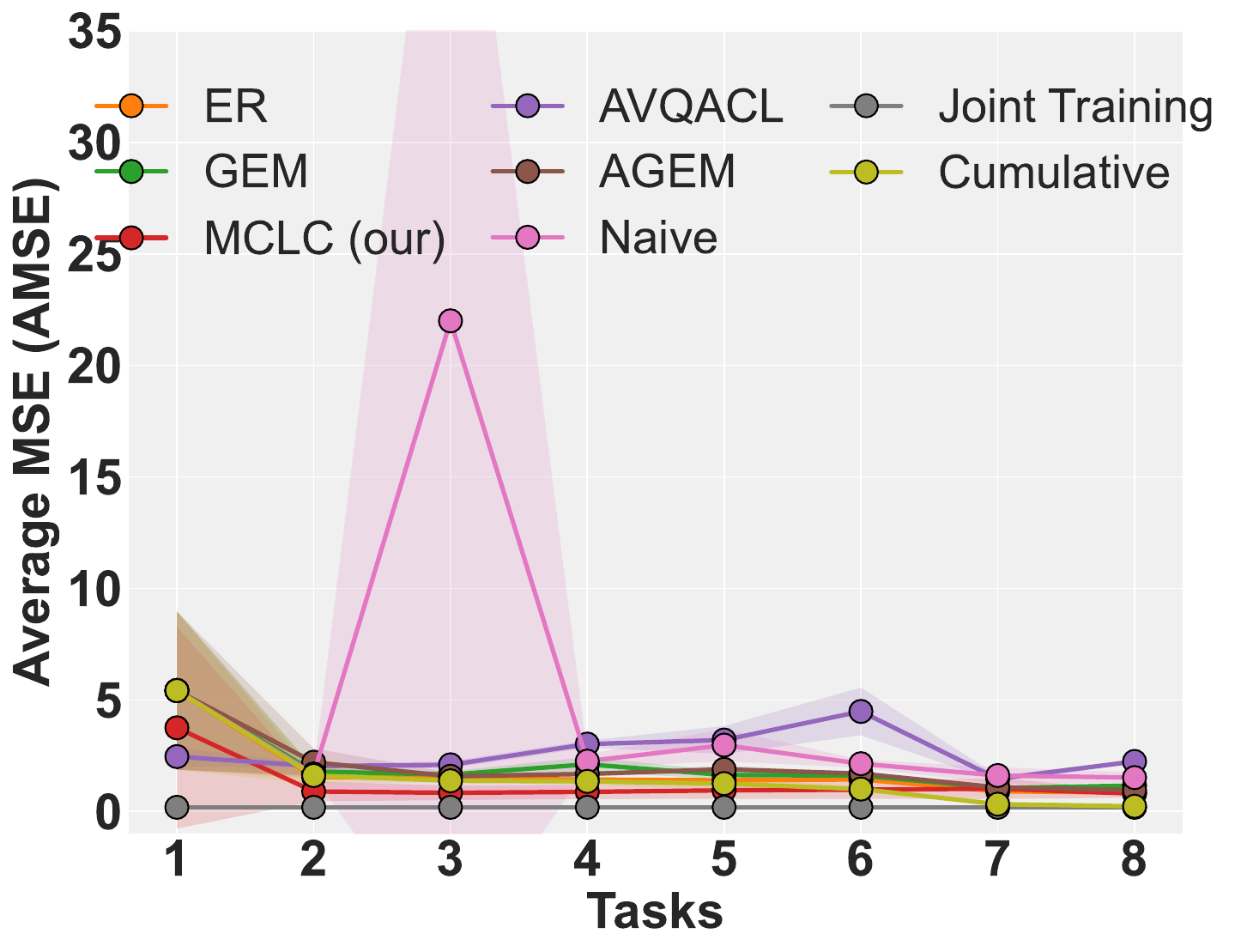} &
\includegraphics[width=0.65\columnwidth]{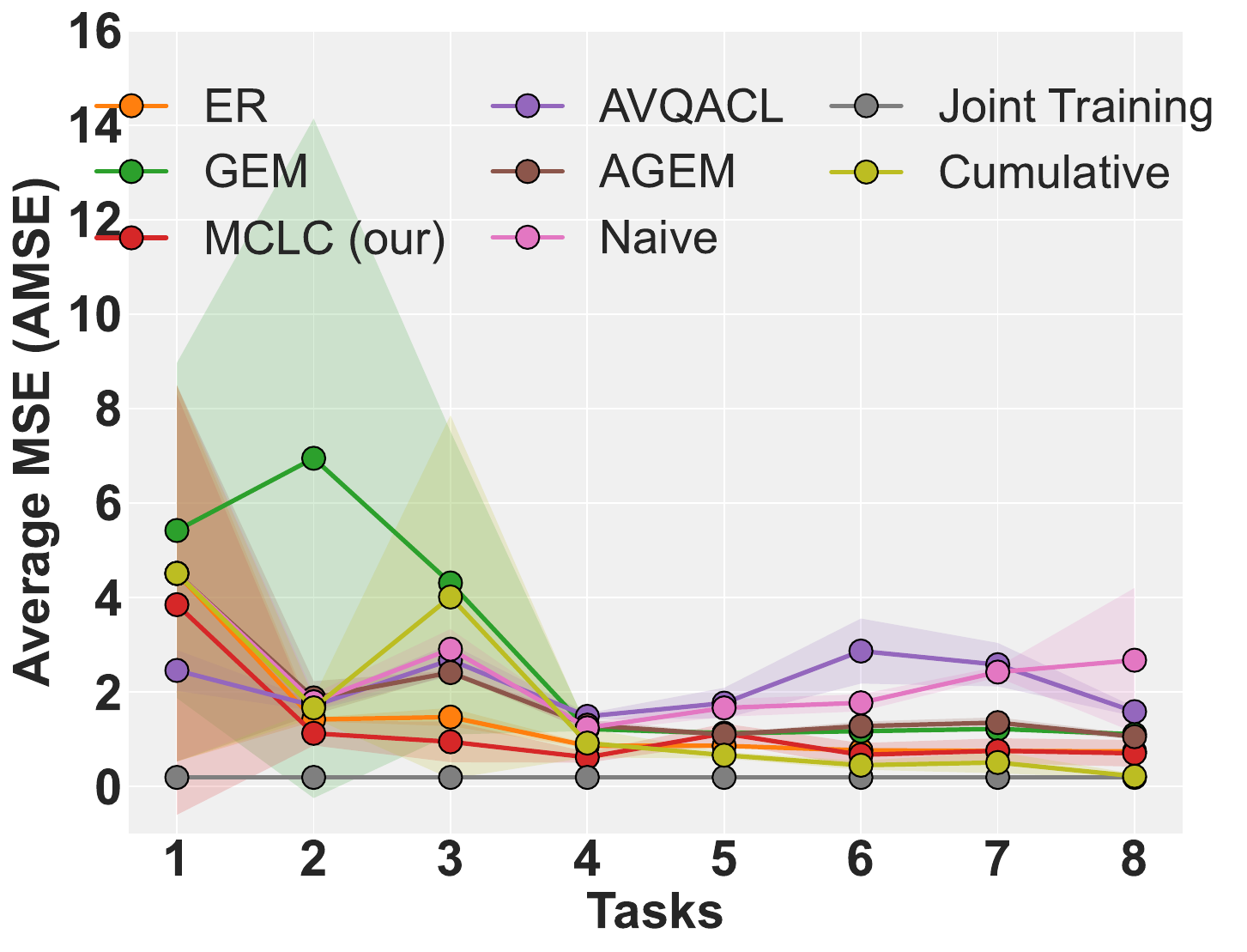} &
\includegraphics[width=0.65\columnwidth]{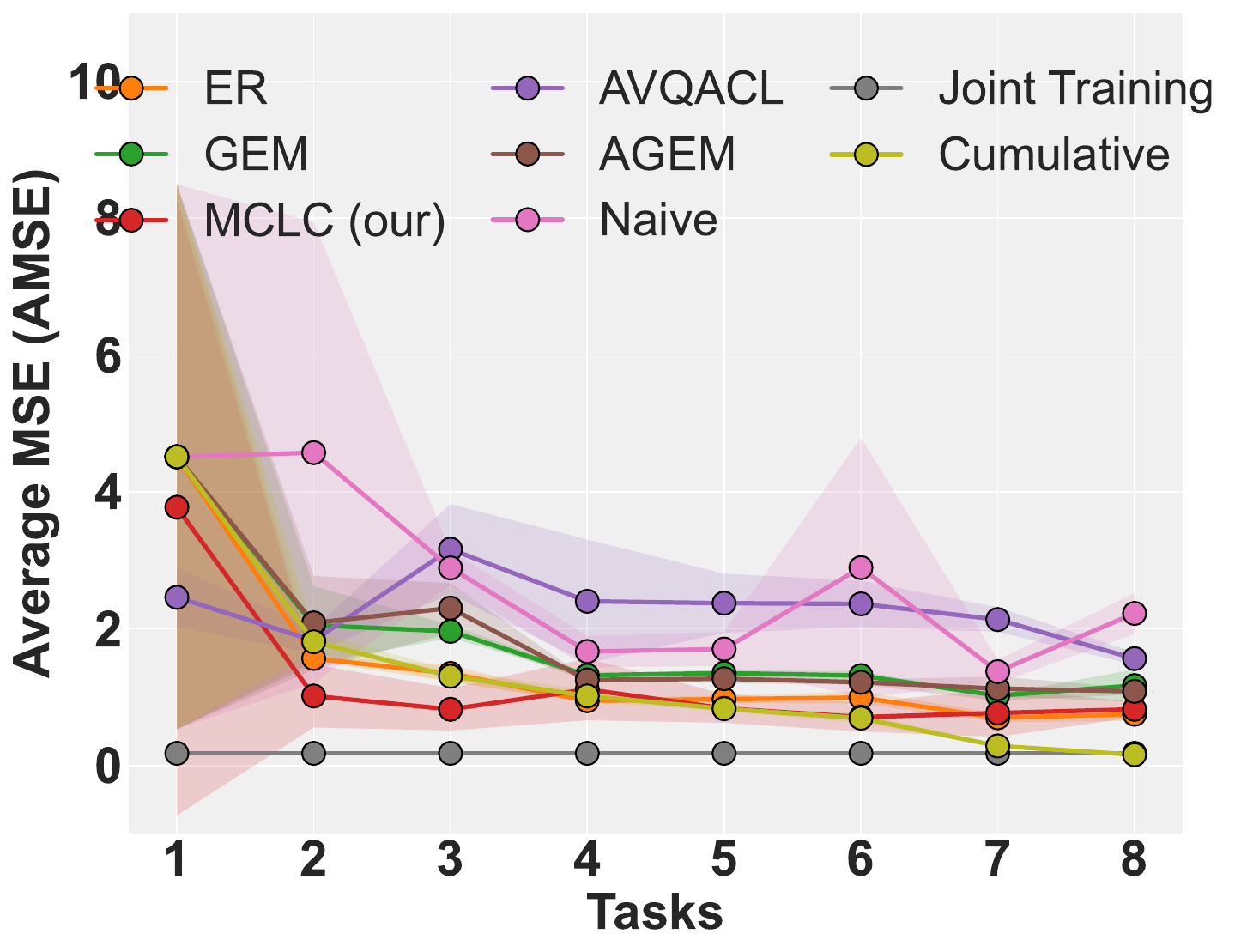} \\

(a) CS1 & (b) CS2 & (c) CS3

\end{tabular}

\caption{AMSE after learning each task for different CL methods under three task sequences of the MMLC dataset. }
\label{fig:mmlc}

\end{figure*}



\begin{figure}
    \centering
    \includegraphics[width=1\linewidth]{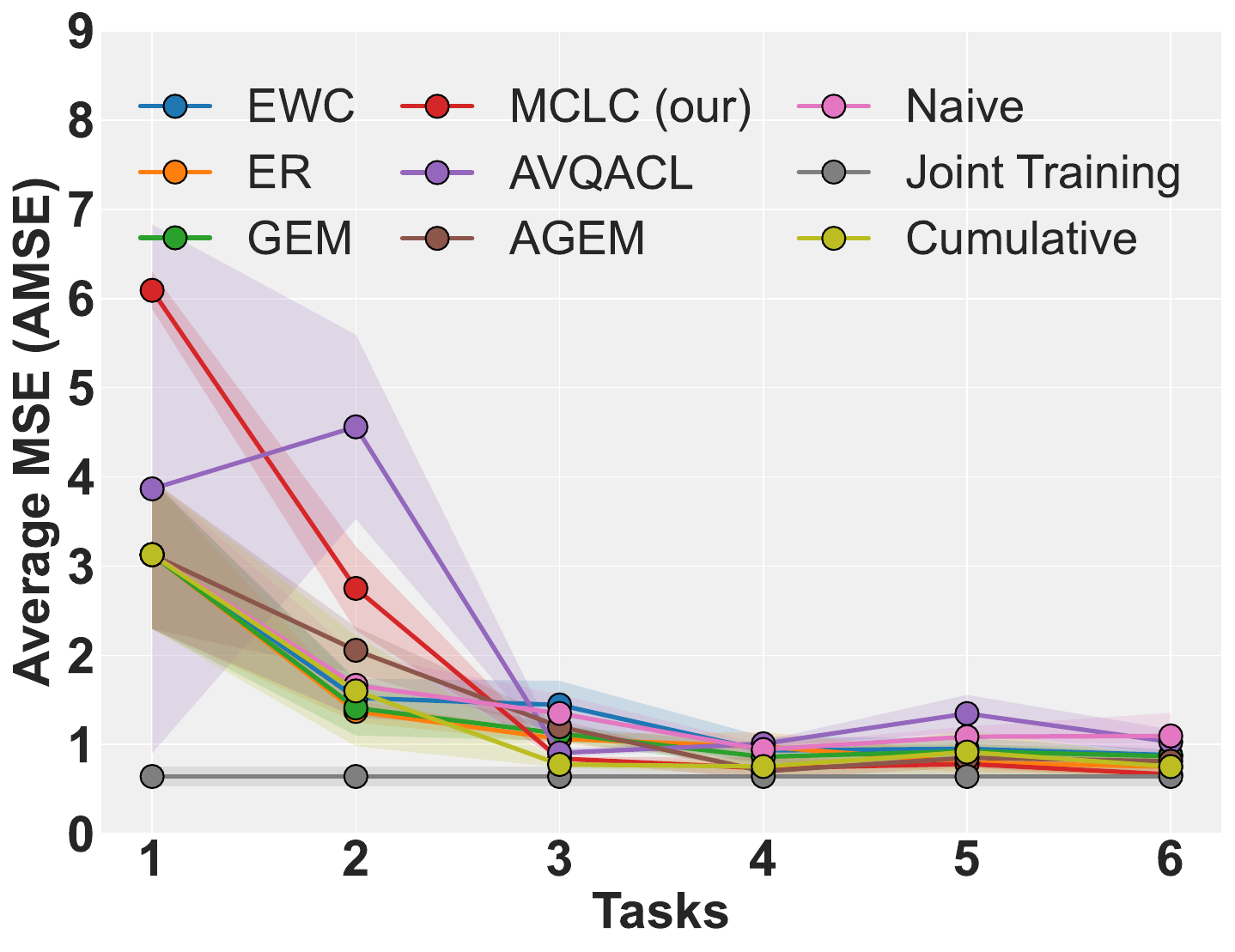}
    \caption{
    AMSE after learning each task on MSU-PID for different CL methods. }
    \label{fig:msupid}
\end{figure}

Figure \ref{fig:mmlc} shows the task-wise AMSE of different methods across the three task sequences of the MMLC dataset. Overall, ER, GEM, A-GEM, and MCLC show a gradual reduction in AMSE as more tasks are learned, indicating improved adaptation in the CL setting. Among the CL methods, MCLC consistently maintains one of the lowest AMSE values across almost all tasks and follows a trend close to the upper-bound methods, Joint Training and Cumulative. ER also demonstrates stable performance but generally exhibits slightly higher errors than MCLC. In contrast, AVQACL and Naive show larger fluctuations across tasks, indicating less stable learning under changing task distributions. Similarly, cumulative and joint training methods maintain consistently low error rates because they have access to all previously seen data during training. For visual clarity, EWC and DER++ are omitted because their larger AMSE values compress the remaining curves. Overall, these results show that MCLC achieves stable learning across different task sequences while effectively reducing catastrophic forgetting. A similar trend is observed in Figure \ref{fig:msupid} on the MSU-PID dataset. Although MCLC starts with a relatively higher AMSE after the first task, its error decreases rapidly and remains consistently low across subsequent tasks. ER also demonstrates stable performance with comparatively low AMSE, whereas AVQACL exhibits larger fluctuations, particularly during the early tasks. GEM and A-GEM perform competitively but generally maintain higher errors than MCLC and ER. As with the MMLC dataset, cumulative and joint training achieve consistently low errors, whereas DER++ is omitted for better visualization. Overall, MCLC demonstrates stable knowledge retention and achieves one of the lowest AMSE values throughout the CL process. 

\subsection{Qualitative result analysis} \label{subsec:qualitative_Analysis}

\begin{figure}[H]
    \centering
    \includegraphics[width=1\linewidth]{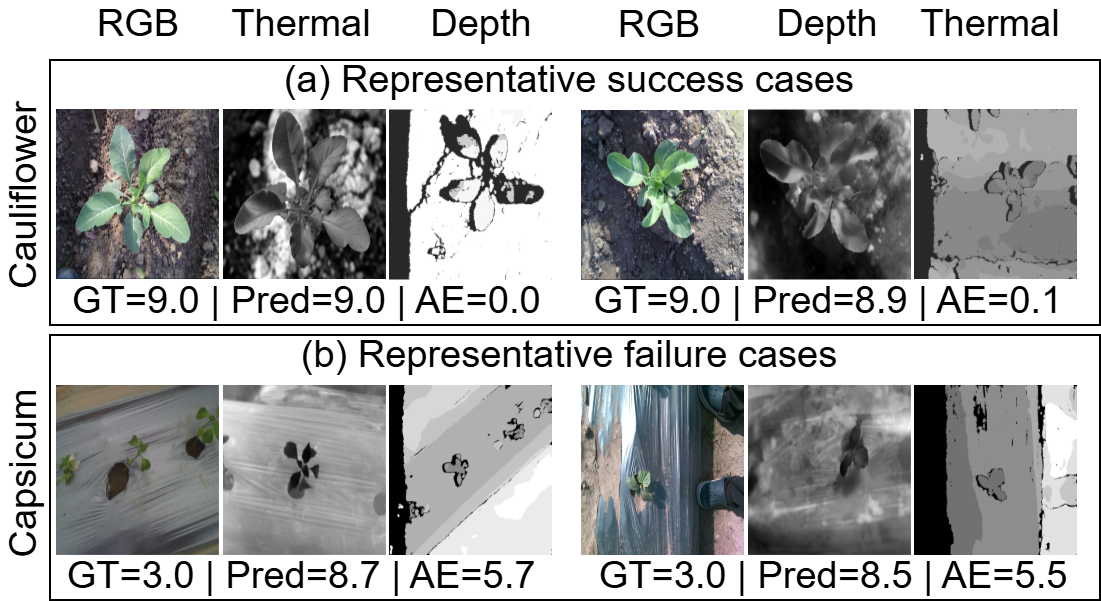}
    \caption{Qualitative analysis of the MCLC framework.}
    \label{fig:qualitative}
\end{figure}

To provide further insight into the behavior of the proposed MCLC framework, Figure \ref{fig:qualitative} presents representative qualitative examples from the test set, including both successful and failure cases. Each example includes the corresponding RGB, thermal, and depth modalities, along with the ground-truth (GT) leaf count, predicted (Pred.) leaf count, and absolute error (AE). As shown in  Figure \ref{fig:qualitative}(a), accurate predictions are obtained when the plant is clearly visible and complementary information is available across all three modalities. The RGB image provides clear visual information, while the thermal and depth images offer complementary cues that help the model estimate the leaf count correctly. As a result, the prediction error is negligible. Figure \ref{fig:qualitative}(b) shows examples where the prediction error is higher. These images contain small plants, uneven illumination, or low-contrast depth and thermal information, making the leaf structure less distinct. Such conditions can lead to an incorrect estimate of the leaf count. Despite these challenging cases, the proposed framework performs reliably on most test samples. 

\section{Ablation Study} \label{ablation}
To comprehensively evaluate the contributions of the various components of our proposed MCLC framework, we conduct a detailed ablation study. This includes (i) comparison among standard leaf counting paradigms to identify the most suitable leaf counting paradigm, (ii) impact of continual multimodal fusion, (iii) performance across different dataset sequences, (iv) effect of memory size and regularization hyperparameters, and (v) performance on other datasets for generalizability.

\subsection{Choice of Leaf Counting Paradigm}
We begin by comparing four standard leaf counting paradigms:  segmentation, regression, density estimation, and object detection on the CVPPP \cite{bell2016aberystwyth} dataset in a non-CL setting. It is a benchmark dataset for leaf counting consisting solely of RGB images. As shown in Table \ref{tab:cvppp}, regression achieves the lowest Train MSE (0.34) and Test MSE (1.08), significantly outperforming others. Hence, regression is chosen as the core paradigm for predicting leaf count in our domain incremental CL setting.

\begin{table}[htbp]
\caption{Performance evaluation of different existing leaf-counting methods on \textbf{the CVPPP dataset}. }
    \centering
    \renewcommand{\arraystretch}{1.1}
    \scalebox{1.1}{
    \begin{tabular}{lcc}
    \hline
        \textbf{Method} & \textbf{Train MSE~$\downarrow$} & \textbf{Test MSE~$\downarrow$}\\ \hline
        Segmentation~\cite{Kuznichov_2019_CVPR_Workshops} & 0.66 & 1.21 \\
        Regression~\cite{giuffrida2018pheno} & \textbf{0.34} & \textbf{1.08} \\
        Density estimation~\cite{aich2018improving} & 23.90 & 21.43 \\
        Object detection~\cite{tu2020automatic} & 16.95 & 62.43 \\ \hline
    \end{tabular}
    }
    \label{tab:cvppp}
\end{table}

\subsection{Effect of Continual Multimodal Fusion}
We evaluated different multimodal fusion methods within the proposed MCLC framework under the domain-incremental CL setting. For a fair comparison, all fusion methods were implemented using the same CL framework. Specifically, we compared our multimodal fusion strategy with \cite{havaei2016hemis}, which performs statistical fusion using the mean and variance of modality-specific features; \cite{shah2023mutex}, a recent transformer-based multimodal learning framework that models cross-modal interactions through attention; and \cite{cai2025keep}, a recent parameter-efficient multimodal framework that employs dynamic sparse cross-modality fusion for feature integration. As shown in Table \ref{tab:fusion}, MCLC consistently outperforms, achieving the lowest BWT (0.432), FWT (0.855), and AMSE (0.861). These results demonstrate that MCLC's continual multimodal fusion strategy is more effective at integrating complementary information from multiple modalities while preserving previously learned knowledge in the CL setting.

\begin{table}[!h]
\caption{Comparison of multimodal fusion methods under the CL setting. }
\centering
\renewcommand{\arraystretch}{1.2}
\scalebox{1.2}{
\begin{tabular}{lcccc}
\hline
\textbf{Method}  & \textbf{BWT~$\downarrow$} & \textbf{FWT~$\downarrow$} & \textbf{AMSE~$\downarrow$} \\
\hline
Havaei et al. \cite{havaei2016hemis} & 0.86 & 2.146 & 1.671 \\ 

Shah et al. \cite{shah2023mutex} &  1.830 & 2.171 & 1.669 \\

Cai et al. \cite{cai2025keep}& 1.449 & 1.632 & 1.256  \\

\hline
MCLC (ours)  & \textbf{0.432} & \textbf{0.855} & \textbf{0.8610}  \\
\hline
\end{tabular}
}
\label{tab:fusion}
\end{table}

\subsection{Impact of Performance across Task Sequences}
To understand how task order influences performance, we evaluated existing methods on the MMLC dataset across three task sequences: CS1, CS2, and CS3. As shown in Table \ref{tab:mmlc_all}, the performance of most CL methods varies with task sequences, demonstrating that task ordering has a significant impact on CL. 

\begin{table*}[htbp]
\caption{Performance comparison of different methods under different hyperparameter (Param.) settings across three task sequences: $CS1$, $CS2$, and $CS3$ of the MMLC dataset. }
\centering
\label{tab:mmlc_all}
\scriptsize                     
\setlength{\tabcolsep}{2.2pt}   
\renewcommand{\arraystretch}{1.15}

\resizebox{\textwidth}{!}{
\begin{tabular}{lc|ccc|ccc|ccc}
\hline
\textbf{Method} & \textbf{Param.}  &
\multicolumn{3}{c|}{\textbf{$CS1$ }} &
\multicolumn{3}{c|}{\textbf{$CS2$ }} & \multicolumn{3}{c}{\textbf{$CS3$ }}\\
& & 
 \textbf{BWT~$\downarrow$} & \textbf{FWT~$\downarrow$} & \textbf{AMSE~$\downarrow$} &
 \textbf{BWT~$\downarrow$} & \textbf{FWT~$\downarrow$} & \textbf{AMSE~$\downarrow$} &
 \textbf{BWT~$\downarrow$} & \textbf{FWT~$\downarrow$} & \textbf{AMSE~$\downarrow$} \\
\hline

\multirow{3}{*}{\begin{tabular}{c}
EWC\\
\cite{kirkpatrick2017overcoming}
\end{tabular}}
& $\lambda=1$
& 2.058$\pm$0.447 & 2.402$\pm$0.602 & 3.110$\pm$1.626
& 4.456$\pm$4.365 & 4.787$\pm$4.459 & 9.002$\pm$12.715
& 2.087$\pm$0.599 & 2.643$\pm$0.928 & 2.596$\pm$0.734 \\

& $\lambda=2$
& 5.592$\pm$6.786 & 10.036$\pm$13.845 & 11.829$\pm$17.133
& 3.675$\pm$2.681 & 3.995$\pm$2.733 & 6.897$\pm$8.877
& 18.935$\pm$29.857 & 19.310$\pm$30.015 & 2.328$\pm$0.237 \\

& $\lambda=3$
& 2.358$\pm$1.026 & 2.699$\pm$0.810 & 4.055$\pm$3.835
& 2.845$\pm$1.244 & 3.255$\pm$1.356 & 4.905$\pm$5.179
& 177.998$\pm$305.667 & 178.405$\pm$305.771 & 2.492$\pm$0.424 \\

\hline

\multirow{4}{*}{\begin{tabular}{c}
GEM\\
\cite{lopez2017gradient}
\end{tabular}}

& $\mathcal{B}$=4$\times$8
&0.794$\pm$0.190 &1.466$\pm$0.364 &1.305$\pm$0.007
& 0.691$\pm$0.376 & 1.567$\pm$0.049 & 1.295$\pm$0.135
&0.662$\pm$0.281 &1.186$\pm$0.030 &1.200$\pm$0.101 \\

& $\mathcal{B}$=5$\times$8
&0.714$\pm$0.308 &1.231$\pm$0.049 &1.221$\pm$0.092
& 0.573$\pm$0.451 & 1.706$\pm$0.332 & 1.399$\pm$0.225
&0.613$\pm$0.179 &1.213$\pm$0.081 &1.202$\pm$0.120 \\

& $\mathcal{B}$=6$\times$8
&0.549$\pm$0.363 &1.205$\pm$0.079 &1.184$\pm$0.033
& 1.012$\pm$0.308 & 1.789$\pm$0.552 & 1.277$\pm$0.042
&0.729$\pm$0.047 &1.149$\pm$0.065 &1.301$\pm$0.040 \\

& $\mathcal{B}$=10$\times$8
&0.647$\pm$0.084 &1.130$\pm$0.080 &1.160$\pm$0.063
& 2.917$\pm$3.656 & 1.398$\pm$0.199 & 1.107$\pm$0.054
&0.680$\pm$0.159 &1.095$\pm$0.068 &1.167$\pm$0.291 \\

\hline

\multirow{4}{*}{\begin{tabular}{c}
A-GEM\\
\cite{chaudhry2018efficient}
\end{tabular}}

& $\mathcal{B}$=4$\times$8
&0.769$\pm$0.300 &1.295$\pm$0.056 &1.071$\pm$0.032
&1.252$\pm$0.114 &1.577$\pm$0.117 &1.130$\pm$0.079
&0.835$\pm$0.105 &1.288$\pm$0.075 &1.301$\pm$0.053 \\

& $\mathcal{B}$=5$\times$8
&0.736$\pm$0.345 &1.278$\pm$0.151 &1.089$\pm$0.106
&1.189$\pm$0.090 &1.544$\pm$0.061 &1.164$\pm$0.016
&0.676$\pm$0.242 &1.217$\pm$0.099 &1.221$\pm$0.103 \\

& $\mathcal{B}$=6$\times$8
&0.692$\pm$0.386 &1.263$\pm$0.091 &1.052$\pm$0.062
&1.057$\pm$0.106 &1.409$\pm$0.103 &1.055$\pm$0.071
&0.774$\pm$0.078 &1.216$\pm$0.091 &1.225$\pm$0.103 \\

& $\mathcal{B}$=10$\times$8
&0.733$\pm$0.217 &1.111$\pm$0.029 &0.954$\pm$0.061
&0.987$\pm$0.035 &1.315$\pm$0.046 &1.077$\pm$0.020
&0.676$\pm$0.056 &1.076$\pm$0.050 &1.082$\pm$0.057 \\
\hline 

\multirow{4}{*}{\begin{tabular}{c}
DER++\\
\cite{buzzega2020dark}
\end{tabular}}

& $\mathcal{B}$=30
& 2.721$\pm$1.485 & 23.220$\pm$2.344 & 27.249$\pm$2.383
& 2.245$\pm$1.257 & 23.960$\pm$1.510 & 26.949$\pm$1.264
& 2.576$\pm$1.823 & 22.832$\pm$1.135 & 27.671$\pm$0.908 \\

& $\mathcal{B}$=40
& 1.576$\pm$1.703 & 21.301$\pm$1.866 & 24.462$\pm$2.159
& 1.959$\pm$0.980 & 23.159$\pm$0.843 & 26.453$\pm$1.327
& 2.695$\pm$0.874 & 21.602$\pm$2.216 & 26.686$\pm$1.731 \\

& $\mathcal{B}$=50
& 1.975$\pm$1.292 & 22.271$\pm$1.694 & 26.995$\pm$1.633
& 1.514$\pm$1.280 & 21.918$\pm$1.572 & 25.591$\pm$1.990
& 1.758$\pm$0.786 & 21.613$\pm$2.484 & 26.132$\pm$2.456 \\

& $\mathcal{B}$=80
& 1.465$\pm$0.787 & 20.386$\pm$1.287 & 25.022$\pm$1.699
& 1.108$\pm$0.463 & 22.119$\pm$0.163 & 25.157$\pm$0.929
& 2.038$\pm$1.432 & 22.004$\pm$2.027 & 26.638$\pm$1.645 \\

\hline

\multirow{4}{*}{\begin{tabular}{c}
ER\\
\cite{rolnick2019experience}
\end{tabular}}

& $\mathcal{B}$=30
& 0.737$\pm$0.079 & 1.106$\pm$0.149 & 1.001$\pm$0.070
& 0.798$\pm$0.135 & 0.993$\pm$0.216 & 1.005$\pm$0.095
& 0.651$\pm$0.100 & 0.898$\pm$0.022 & 1.052$\pm$0.104 \\

& $\mathcal{B}$=40
& 0.735$\pm$0.136 & 0.949$\pm$0.157 & 0.996$\pm$0.165
& 0.765$\pm$0.094 & 0.997$\pm$0.095 & 0.898$\pm$0.062
& 0.503$\pm$0.088 & 0.773$\pm$0.056 & 0.945$\pm$0.035 \\

& $\mathcal{B}$=50
& 0.630$\pm$0.153 & 0.927$\pm$0.060 & 0.899$\pm$0.012
& 0.714$\pm$0.067 & 0.941$\pm$0.092 & 0.863$\pm$0.045
& 0.467$\pm$0.172 & 0.745$\pm$0.064 & 0.894$\pm$0.093 \\

& $\mathcal{B}$=80
& 0.565$\pm$0.144 & 0.826$\pm$0.056 & 0.843$\pm$0.090
& 0.564$\pm$0.069 & 0.790$\pm$0.031 & 0.743$\pm$0.049
& 0.394$\pm$0.038 & 0.624$\pm$0.061 & 0.749$\pm$0.120 \\

\hline

\multirow{4}{*}{\begin{tabular}{c}
AVQACL\\
\cite{wu2025avqacl}
\end{tabular}}

& $\mathcal{B}$=30
& 1.788$\pm$0.536 & 2.516$\pm$0.495 & 2.726$\pm$1.421
& 1.641$\pm$0.267 & 2.354$\pm$0.261 & 1.889$\pm$0.145
& 1.198$\pm$0.201 & 1.893$\pm$0.136 & 1.617$\pm$0.094 \\

& $\mathcal{B}$=40
& 1.401$\pm$0.172 & 2.114$\pm$0.152 & 2.565$\pm$0.631
& 1.692$\pm$0.246 & 2.445$\pm$0.217 & 1.843$\pm$0.171
& 1.325$\pm$0.267 & 1.979$\pm$0.236 & 1.561$\pm$0.115 \\

& $\mathcal{B}$=50
& 1.368$\pm$0.300 & 2.084$\pm$0.281 & 2.241$\pm$0.114
& 1.422$\pm$0.145 & 2.154$\pm$0.121 & 1.585$\pm$0.128
& 1.272$\pm$0.305 & 1.963$\pm$0.300 & 1.748$\pm$0.286 \\

& $\mathcal{B}$=80
& 2.018$\pm$0.847 & 2.681$\pm$0.962 & 2.629$\pm$0.635
& 1.655$\pm$0.381 & 2.405$\pm$0.364 & 1.780$\pm$0.083
& 1.375$\pm$0.304 & 2.050$\pm$0.311 & 1.933$\pm$0.510 \\

\hline

\multirow{4}{*}{
\begin{tabular}{c}
\textbf{MCLC}\\
\textbf{(Ours)}
\end{tabular}
}

& $\mathcal{B}$=30
& 0.334$\pm$0.090 & 0.772$\pm$0.176 & 1.376$\pm$0.477
& 0.192$\pm$0.108 & 0.769$\pm$0.018 & 0.736$\pm$0.176
& 0.097$\pm$0.020 & 0.686$\pm$0.080 & 1.063$\pm$0.235 \\

& $\mathcal{B}$=40
& 0.218$\pm$0.071 & 0.654$\pm$0.157 & 0.972$\pm$0.272
& 0.145$\pm$0.163 & 0.756$\pm$0.111 & 0.742$\pm$0.213
& 0.258$\pm$0.219 & 0.714$\pm$0.132 & 1.084$\pm$0.145 \\

& $\mathcal{B}$=50
& \textbf{0.186$\pm$0.083} & \textbf{0.557$\pm$0.063} & 0.795$\pm$0.201
& 0.174$\pm$0.038 & 0.709$\pm$0.052 & 0.655$\pm$0.088
& 0.297$\pm$0.201 & 0.721$\pm$0.143 & 1.066$\pm$0.308 \\

& $\mathcal{B}$=80
& 0.280$\pm$0.115 & 0.661$\pm$0.104 & \textbf{0.675$\pm$0.027}
& \textbf{0.145$\pm$0.116} & \textbf{0.644$\pm$0.037} & \textbf{0.542$\pm$0.069}
&\textbf{ 0.055$\pm$0.038} & \textbf{0.594$\pm$0.050} & \textbf{0.745$\pm$0.057} \\

\hline

\end{tabular}}
\end{table*}

In $CS1$, tasks are grouped by crop type, resulting in gradual domain transitions. This enables relatively stable learning for most methods. For example, ER achieves an AMSE of 0.843$\pm$0.090, while the MCLC further reduces the error to 0.675$\pm$0.027 using a buffer size of 80. Regularization-based EWC, however, exhibits considerably higher errors, with the best AMSE of 3.110$\pm$1.626. In $CS2$, tasks are organized by image capture time, thereby introducing greater variation across crops. This sequence is more challenging for most methods, as reflected in the substantial performance degradation of EWC, whose AMSE increases to 9.002$\pm$12.715 even with its best-performing hyperparameter. ER remains relatively robust with an AMSE of 0.743$\pm$0.049, whereas the proposed MCLC achieves the lowest AMSE of 0.542$\pm$0.069, together with the lowest BWT (0.145$\pm$0.116) and FWT (0.644$\pm$0.037), demonstrating improved knowledge retention and forward transfer. In $CS3$, tasks are arranged in a non-linear mixed order that combines different crops and capture times. This sequence, therefore, introduces abrupt distributional changes. Although this sequence causes noticeable performance fluctuations for several methods, the proposed MCLC continues to achieve the best overall performance with an AMSE of 0.745$\pm$0.057, outperforming ER (0.749$\pm$0.120) and AVQACL (1.561$\pm$0.115). In contrast, EWC exhibits extremely high variability, with BWT and FWT values of 18.935$\pm$29.857 and 19.310$\pm$30.015, indicating high sensitivity to task ordering. 

From these observations, we conclude that task sequencing significantly influences CL performance. Also, regularization-based methods such as EWC are highly sensitive to task order and distributional shifts. Moreover, rehearsal-based methods provide greater stability and consistently lower errors across sequences. More importantly, the proposed MCLC consistently achieves the lowest AMSE across all three sequences while maintaining low BWT and FWT values, demonstrating strong robustness, effective knowledge retention, and improved generalization under diverse multimodal CL scenarios.

\subsection{Impact of Buffer and Hyperparameters}

Buffer size and hyperparameters have a clear impact on performance across all sequences, as shown in Table \ref{tab:mmlc_all}. Regularization-based methods such as EWC rely on the regularization parameter $\lambda$ to preserve previously learned knowledge. However, EWC exhibits inconsistent performance across different values of $\lambda$, with large variations in both the mean and standard deviation. In particular, increasing $\lambda$ does not consistently improve performance. This indicates that simply strengthening the regularization term is insufficient to effectively prevent catastrophic forgetting under large distributional changes. For memory-based methods, the buffer memory plays an important role in improving performance. As the buffer memory increases from $4\times8$ to $10\times8$, both GEM and A-GEM generally achieve lower AMSE values. For example, GEM achieves its best performance with a buffer size of $10\times8$, obtaining AMSE values of 1.160$\pm$0.063 in CS1, 1.107$\pm$0.054 in CS2, and 1.167$\pm$0.291 in CS3. Similarly, A-GEM also benefits from larger buffer memory, obtaining its lowest AMSE values of 0.954$\pm$0.061, 1.055$\pm$0.071, and 1.082$\pm$0.057 across CS1, CS2, and CS3, respectively. These results suggest that increasing buffer memory helps reduce forgetting, although the performance improvement gradually saturates. 

ER shows the clearest effect of buffer size. As the buffer memory increases from 30 to 80, the AMSE consistently decreases across all task sequences. For instance, the AMSE improves from 1.001$\pm$0.070 to 0.843$\pm$0.090 in CS1, from 1.005$\pm$0.095 to 0.743$\pm$0.049 in CS2, and from 1.052$\pm$0.104 to 0.749$\pm$0.120 in CS3. Similarly, AVQACL benefits from larger buffer sizes and achieves its best performance with more memory, although its comparatively large standard deviations indicate greater sensitivity to initialization across different random seeds. Compared with all existing methods, the proposed MCLC consistently achieves the best performance across different buffer settings. Even with moderate buffer sizes of 30 and 40, MCLC performs competitively, and increasing the buffer size to 80 further improves performance, achieving the lowest AMSE values of 0.675$\pm$0.027, 0.542$\pm$0.069, and 0.745$\pm$0.057 across CS1, CS2, and CS3, respectively. Moreover, MCLC also achieves the lowest BWT and FWT values in most settings. 

Overall, these results show that memory plays a crucial role in CL, and increasing buffer size generally improves performance. More importantly, the way memory is utilized is equally important. The proposed MCLC method demonstrates that selecting high-quality, representative samples yields better performance than simply increasing memory size.

\subsection{Evaluating Generalization Across Datasets}

To further validate the effectiveness of our approach, we conducted experiments on an additional dataset, MSU-PID, using the same set of CL methods. As shown in Table \ref{tab:msupid_data}, the results follow trends similar to those observed on the MMLC dataset. Rehearsal-based methods clearly benefit from increased buffer sizes. For example, ER shows a consistent reduction in AMSE as the buffer size increases, improving from 0.968$\pm$0.009 ($\mathcal{B}$=30) to 0.750$\pm$0.109 ($\mathcal{B}$=80). Similarly, GEM and A-GEM also achieve lower AMSE values with larger buffers, reaching their best performance of 0.869$\pm$0.052 and 0.813$\pm$0.084, respectively, at $\mathcal{B}$=10$\times$8. However, regularization-based methods like EWC still perform relatively worse, with AMSE remaining between 0.882$\pm$0.074 and 0.896$\pm$0.135, highlighting their limitations in handling changing data distributions. DER++ performs significantly worse, with very high AMSE values, indicating instability in this domain incremental CL setting. AVQACL shows competitive performance at smaller buffer sizes but does not improve consistently as buffer size increases. Notably, our proposed MCLC outperforms all baselines, achieving the lowest AMSE of 0.664$\pm$0.071 ($\mathcal{B}$=80) along with the lowest FWT of 0.979$\pm$0.104. These results indicate that MCLC effectively preserves prior knowledge while adapting to new tasks, making it a robust and scalable solution across diverse datasets.

\begin{table}[htbp!]
\caption{Performance comparison of different approaches under various hyperparameter settings on the MSU-PID dataset.}
\centering
\label{tab:msupid_data}
\renewcommand{\arraystretch}{1.3}
\scalebox{0.75}{
\begin{tabular}{lc|ccc}
\hline
\textbf{Approach} & \textbf{Parameter} &
\textbf{BWT $\downarrow$} &
\textbf{FWT $\downarrow$} &
\textbf{AMSE $\downarrow$} \\
\hline

\multirow{3}{*}{\begin{tabular}{c}
EWC\\
\cite{kirkpatrick2017overcoming}
\end{tabular}}
& $\lambda$=1 & 1.029 $\pm$ 0.514 & 1.246 $\pm$ 0.118 & 0.883 $\pm$ 0.079 \\
& $\lambda$=2 & 1.062 $\pm$ 0.575 & 1.203 $\pm$ 0.105 & 0.882 $\pm$ 0.074 \\
& $\lambda$=3 & 1.082 $\pm$ 0.553 & 1.245 $\pm$ 0.045 & 0.896 $\pm$ 0.135 \\
\hline

\multirow{4}{*}{\begin{tabular}{c}
GEM\\
\cite{lopez2017gradient}
\end{tabular}}
& $\mathcal{B}=$4x8 & 0.918 $\pm$ 0.455 & 1.405 $\pm$ 0.141 & 1.020 $\pm$ 0.169 \\
& $\mathcal{B}=$5x8 & 1.190 $\pm$ 0.652 & 1.202 $\pm$ 0.024 & 1.058 $\pm$ 0.154 \\
& $\mathcal{B}=$6x8 & 0.975 $\pm$ 0.421 & 1.218 $\pm$ 0.094 & 0.996 $\pm$ 0.042 \\
& $\mathcal{B}=$10x8 & 1.134 $\pm$ 0.356 & 1.103 $\pm$ 0.067 & 0.869 $\pm$ 0.052 \\
\hline

\multirow{4}{*}{\begin{tabular}{c}
A-GEM\\
\cite{chaudhry2018efficient}
\end{tabular}}
& $\mathcal{B}=$4x8 & 0.984 $\pm$ 0.437 & 1.234 $\pm$ 0.059 & 0.920 $\pm$ 0.066 \\
& $\mathcal{B}=$5x8 & 1.009 $\pm$ 0.541 & 1.121 $\pm$ 0.103 & 0.849 $\pm$ 0.082 \\
& $\mathcal{B}=$6x8 & 1.135 $\pm$ 0.665 & 1.125 $\pm$ 0.164 & 0.859 $\pm$ 0.016 \\
& $\mathcal{B}=$10x8 & 1.073 $\pm$ 0.574 & 1.116 $\pm$ 0.098 & 0.813 $\pm$ 0.084 \\
\hline

\multirow{4}{*}{\begin{tabular}{c}
DER++\\
\cite{buzzega2020dark}
\end{tabular}}
& $\mathcal{B}$=30 & -5.514 $\pm$ 19.024 & 44.641 $\pm$ 3.120 & 49.140 $\pm$ 4.562 \\
& $\mathcal{B}$=40 & -4.497 $\pm$ 14.004 & 37.022 $\pm$ 3.555 & 37.943 $\pm$ 4.602 \\
& $\mathcal{B}$=50 & -3.888 $\pm$ 17.017 & 38.897 $\pm$ 4.775 & 41.953 $\pm$ 5.590 \\
& $\mathcal{B}$=80 & -3.656 $\pm$ 14.338 & 36.909 $\pm$ 6.826 & 37.616 $\pm$ 6.954 \\
\hline

\multirow{4}{*}{\begin{tabular}{c}
ER\\
\cite{rolnick2019experience}
\end{tabular}}
& $\mathcal{B}$=30 & 0.986 $\pm$ 0.347 & 1.139 $\pm$ 0.050 & 0.968 $\pm$ 0.009 \\
& $\mathcal{B}$=40 & 1.031 $\pm$ 0.383 & 1.050 $\pm$ 0.022 & 0.857 $\pm$ 0.099 \\
& $\mathcal{B}$=50 & 1.119 $\pm$ 0.564 & 1.046 $\pm$ 0.064 & 0.862 $\pm$ 0.032 \\
& $\mathcal{B}$=80 & 1.144 $\pm$ 0.590 & 1.031 $\pm$ 0.133 & 0.750 $\pm$ 0.109 \\
\hline

\multirow{4}{*}{\begin{tabular}{c}
AVQACL\\
\cite{wu2025avqacl}
\end{tabular}}
& $\mathcal{B}$=30 & 0.104 $\pm$ 0.467 & 1.385 $\pm$ 0.333 & 1.180 $\pm$ 0.136 \\
& $\mathcal{B}$=40 & 0.060 $\pm$ 0.630 & 1.336 $\pm$ 0.357 & 1.277 $\pm$ 0.380 \\
& $\mathcal{B}$=50 & 0.104 $\pm$ 0.565 & 1.345 $\pm$ 0.311 & 1.349 $\pm$ 0.451 \\
& $\mathcal{B}$=80 & 0.095 $\pm$ 0.292 & 1.183 $\pm$ 0.186 & 1.030 $\pm$ 0.170 \\
\hline

\multirow{4}{*}{\begin{tabular}{c}
\textbf{MCLC}\\
\textbf{(Ours)}
\end{tabular}}
& $\mathcal{B}$=30 & 2.113 $\pm$ 0.248 & 1.099 $\pm$ 0.081 & 0.936 $\pm$ 0.161 \\
& $\mathcal{B}$=40 & 2.159 $\pm$ 0.188 & 1.071 $\pm$ 0.055 & 0.864 $\pm$ 0.006 \\
& $\mathcal{B}$=50 & 2.154 $\pm$ 0.201 & 1.053 $\pm$ 0.053 & 0.798 $\pm$ 0.028 \\
& \textbf{$\mathcal{B}$=80} & $\mathbf{2.244 \pm 0.148}$ & $\mathbf{0.979 \pm 0.104}$ & $\mathbf{0.664 \pm 0.071}$ \\
\hline

\end{tabular}
}
\end{table}
\balance
\section{Conclusion} \label{conclusion}


We address leaf counting in dynamic agricultural environments using a CL framework with multimodal data. We propose MCLC, which combines cross-modal fusion with a replay-based memory to learn tasks sequentially while reducing forgetting. Unlike traditional methods, MCLC can adapt to changing data distributions while retaining previously learned knowledge. The experimental results, averaged over three random seeds, show that MCLC consistently outperforms the baseline methods across all data sequences.For instance, in CS1, it achieves the lowest AMSE of 0.675$\pm$0.027, while in CS2 and CS3 it further achieves AMSE values of 0.542$\pm$0.069 and 0.745$\pm$0.057, respectively. It also maintains low BWT and FWT values, indicating effective knowledge retention and forward transfer across different task orders. Compared with EWC and memory-based methods such as ER, GEM and A-GEM, MCLC consistently achieves better performance, particularly with larger buffer sizes. Overall, MCLC provides a reliable and scalable solution for multimodal continual leaf counting. Its ability to handle distribution changes, reduce forgetting, and make effective use of memory makes it useful not only for plant phenotyping but also for other real-world CL applications.



\section{Limitations and Future Work} \label{futurework}

Although the proposed MCLC framework demonstrates strong performance for multimodal continual leaf counting, it assumes the availability of aligned RGB, depth, and thermal modalities during both training and inference. In practical agricultural deployments, some modalities may be missing, noisy, or affected by sensor failures, environmental conditions, or unavailability. Such scenarios may reduce the effectiveness of multimodal fusion and continual adaptation. Also, MCLC currently relies on a fixed memory buffer, which may be challenging for large-scale, long-term CL settings. 

Future work will focus on developing more robust multimodal CL strategies that handle missing modalities and noisy sensor inputs and on more memory-efficient buffer mechanisms for real-world agricultural applications.

\section*{Acknowledgments}
The authors acknowledge ANNAM.AI, an AI-CoE of the Ministry of Education, Govt. of India, at the Indian Institute of Technology Ropar, for providing resources and support to execute this work. This work is also supported by a grant from DST, Govt. of India, for the Technology Innovation Hub at IIT Ropar in the framework of the National Mission on Interdisciplinary Cyber-Physical Systems.

\bibliographystyle{IEEEtran}
 \bibliography{bibfile}

\end{document}